\documentclass[runningheads]{llncs}
\usepackage[T1]{fontenc}
\usepackage{graphicx}
\usepackage{amsmath}
\usepackage{amsfonts}
\usepackage{array}
\usepackage{tikz}
\usetikzlibrary{shapes.geometric, arrows, graphs, positioning, quotes, fit, decorations.pathreplacing}
\usepackage{rotating}
\graphicspath{{images/}}
\begin{document}
\title{REVNET: Rotation-Equivariant Point Cloud Completion via Vector Neuron
Anchor Transformer}
\titlerunning{Rotation-Equivariant Point Cloud Completion via VN
Anchor Transformer}
%
\author{Zhifan Ni\inst{1} \and
Eckehard Steinbach\inst{1}}
\authorrunning{Z. Ni and E. Steinbach}
%
\institute{Technical University of Munich, Munich, Germany \\
\email{\{zhifan.ni, eckehard.steinbach\}@tum.de}}
\maketitle              
%

%
%
%


\begin{abstract}
Incomplete point clouds captured by 3D sensors often result in the loss of both geometric and semantic information. Most existing point cloud completion methods are built on rotation-variant frameworks trained with data in canonical poses, limiting their applicability in real-world scenarios. While data augmentation with random rotations can partially mitigate this issue, it significantly increases the learning burden and still fails to guarantee robust performance under arbitrary poses. To address this challenge, we propose the Rotation-Equivariant Anchor Transformer (REVNET), a novel framework built upon the Vector Neuron (VN) network for robust point cloud completion under arbitrary rotations. To preserve local details, we represent partial point clouds as sets of equivariant anchors and design a VN Missing Anchor Transformer to predict the positions and features of missing anchors. Furthermore, we extend VN networks with a rotation-equivariant bias formulation and a ZCA-based layer normalization to improve feature expressiveness. Leveraging the flexible conversion between equivariant and invariant VN features, our model can generate point coordinates with greater stability. Experimental results show that our method outperforms state-of-the-art approaches on the synthetic MVP dataset in the equivariant setting. On the real-world KITTI dataset, REVNET delivers competitive results compared to non-equivariant networks, without requiring input pose alignment. The source code will be released on GitHub under URL: \url{https://github.com/nizhf/REVNET}.

\end{abstract}
    
\section{Introduction}
\label{sec:intro}
3D sensors are widely used to capture geometric information of real-world scenes. However, the resulting point clouds are often incomplete due to occlusion and sensor limitations, adversely affecting downstream tasks like shape retrieval~\cite{3d_retrieval:deepspf} and point cloud registration~\cite{pc_regist:hegn}. This makes point cloud completion, recovering a complete shape from a partial observation, a crucial task in 3D vision. 

\begin{figure}
    \centering
    \begin{tikzpicture}[font=\footnotesize, >={latex}]
        \node[inner sep=0pt] (part1) {\includegraphics[height=42pt]{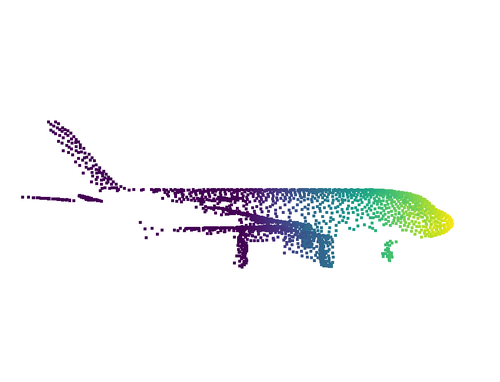}};
        \node[inner sep=0pt, right=20pt of part1] (comp1) {\includegraphics[height=42pt]{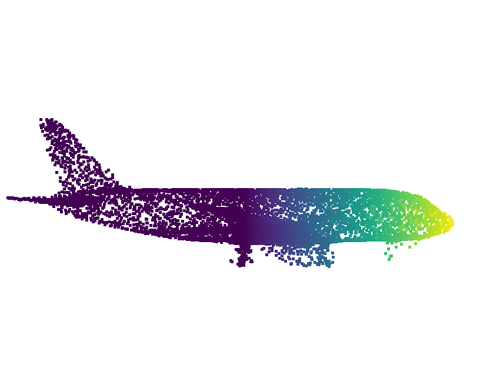}};
        \node[inner sep=0pt, right=30pt of comp1] (part1r) {\includegraphics[height=42pt]{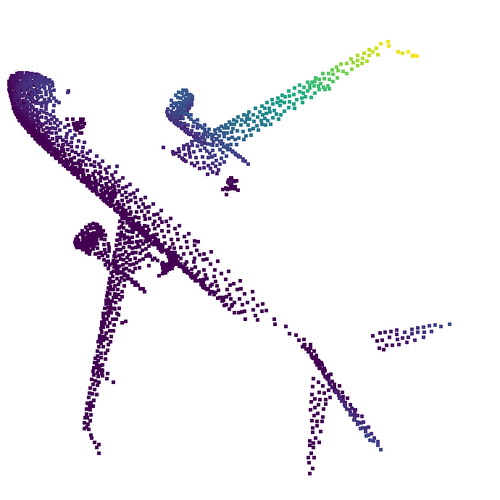}};
        \node[inner sep=0pt, right=30pt of part1r] (comp1r) {\includegraphics[height=42pt]{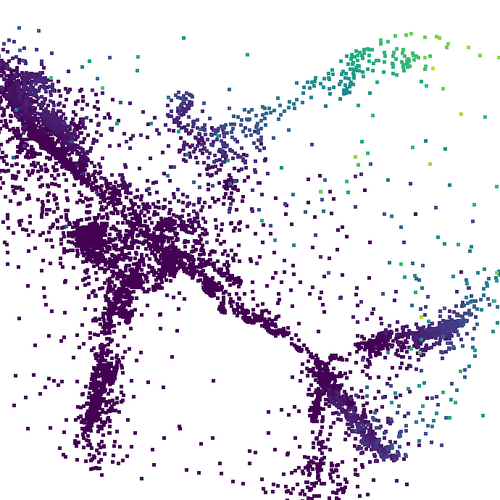}};

        \node[draw=none, rectangle, align=center, above=-7pt of part1, xshift=110pt] {(a) Non-equivariant Methods};

        \node[inner sep=0pt, below=25pt of part1] (part2) {\includegraphics[height=42pt]{intro/24352_partial.png}};
        \node[inner sep=0pt, right=20pt of part2] (comp2) {\includegraphics[height=42pt]{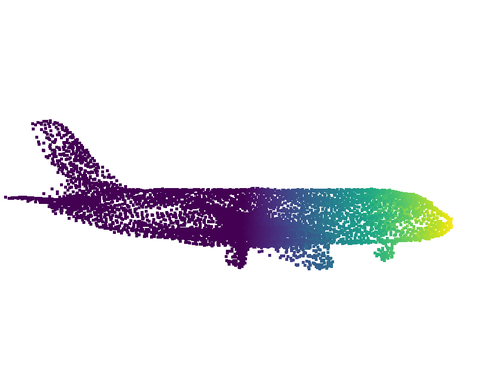}};
        \node[inner sep=0pt, right=30pt of comp2] (part2r) {\includegraphics[height=42pt]{intro/24352r_partial.png}};
        \node[inner sep=0pt, right=30pt of part2r] (comp2r) {\includegraphics[height=42pt]{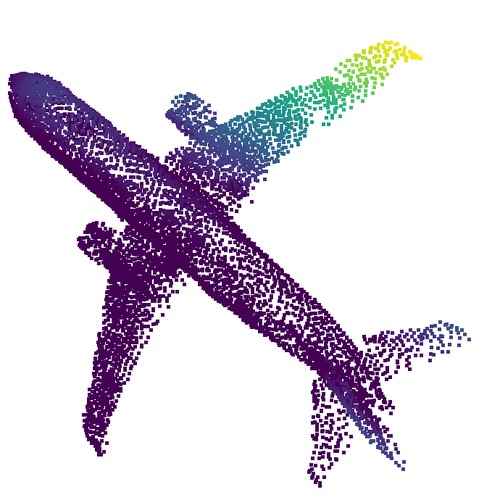}};

        \node[draw=none, rectangle, align=center, above=-7pt of part2, xshift=110pt] {(b) REVNET};

        \node[draw=none, rectangle, align=center, below=3pt of part1] (cano) {Canonical Pose};
        \node[draw=none, rectangle, align=center, below=3pt of comp1] (cano_comp) {Completion};
        \node[draw=none, rectangle, align=center, below=3pt of part1r] (arbi) {Arbitrary Pose};
        \node[draw=none, rectangle, align=center, below=3pt of comp1r] (arbi_comp) {Completion};

        \draw[->] (part1.east) -> (comp1.west);
        \draw[->] (part1r.east) -> ([xshift=-5pt]comp1r.west);
        \draw[->] (part2.east) -> (comp2.west);
        \draw[->] (part2r.east) -> ([xshift=-5pt]comp2r.west);
        \draw[->] (cano.east) -> ([yshift=0.8pt]cano_comp.west);
        \draw[->] (arbi.east) -> (arbi_comp.west);

        \node[draw, dashed, rectangle, align=center, minimum height=52pt, minimum width=300pt, above right=-44pt and -65pt of part1] (box_a) {};
        \node[draw, dashed, rectangle, align=center, minimum height=52pt, minimum width=300pt, below=15pt of box_a] (box_b) {};

    \end{tikzpicture}
    \caption{(a) Conventional point cloud completion methods are trained with data in canonical poses and cannot handle observation under pose changes. (b) Our proposed REVNET can consistently recover shapes under arbitrary poses.}
    \label{fig:intro}
\end{figure}

Recent point cloud completion networks have evolved from global-encoding paradigms~\cite{comp:pcn,comp:topnet,comp:vrcnet,comp:snowflakenet} to anchor-based architectures~\cite{comp:pointr,comp:anchorformer,comp:adapointr,comp:odgnet} that predict missing structures from local keypoint features, leading to improved geometric fidelity. Despite these advances, most existing methods still assume that inputs are pre-aligned to a canonical pose, posing significant challenges for real-world deployment. As shown in Fig.~\ref{fig:intro}, variations in input orientation can cause severe performance degradation. To mitigate this, prior works rely on pose alignment via pose estimation~\cite{comp_equiv:scarp}, which is error-prone and may further degrade the quality. Alternatively, applying rotation augmentation during training increases the learning burden and often leads to instability~\cite{comp_equiv:equiv_pcn}. To address these challenges, the task of SO(3)-equivariant point cloud completion aims to generate consistent and complete 3D shapes from partial observations under arbitrary rotations, even when trained solely on aligned data. A pioneering attempt, EquivPCN~\cite{comp_equiv:equiv_pcn}, utilizes the rotation-equivariant Vector Neuron (VN) framework~\cite{rotequiv:vnn}. However, EquivPCN follows the early paradigm based on global feature encoding, where critical local information is often lost due to pooling operations.

In this work, we propose the Rotation-Equivariant VN Anchor Transformer (REVNAT), a novel framework that overcomes this limitation by encoding the input into a set of VN anchor features, and inferring the missing regions via a Transformer-based architecture tailored for VN representations. 
The main contributions of our work can be summarized as: (1) We propose the rotation-equivariant Anchor Transformer, which adopts multi-head channel-wise subtraction attention to robustly estimate missing anchor features. (2) We leverage the flexible equivariant-invariant conversion property of VN to enable stable coordinates prediction. (3) We develop a VN-based feature backbone with an equivariant bias formulation and a layer normalization based on zero-phase component analysis (ZCA) to improve the overall VN model performance.


\section{Related Work}
\label{sec:related}

\subsubsection{Rotation-Aware Point Cloud Analysis.}

Following the pioneering PointNet~\cite{3d_point:pointnet}, numerous learning-based point cloud networks~\cite{3d_point:pointnet++,3d_point:dgcnn,3d_point:pointmetabase,3d_point:pt,3d_point:pointcnn} have been proposed for 3D tasks such as object recognition and semantic segmentation. However, these frameworks are not rotation-aware: the same shape in different orientations yields inconsistent latent representations. To address this, rotation-invariant approaches encode input geometry into pose-independent features using hand-crafted descriptors~\cite{rotinv:riconv++,rotinv:crin,rotinv:sgmnet} or PCA-based alignment~\cite{rotinv:endowing_deep,rotinv:closer_look}. However, such methods cannot model how features evolve under rotation, making them unsuitable for equivariant shape completion. Alternatively, rotation-equivariant networks explicitly model how features transform with input rotations. Tensor field networks~\cite{rotequiv:tfn_nonlinear} leverage spherical harmonics, and spherical CNNs~\cite{rotequiv:spherical_cnn,rotequiv:sfcnn} operate on projected spherical grids. However, their computational overhead limits their scalability and practical application. More efficient alternatives represent features as quaternions~\cite{rotequiv:reqnn} or vector sets~\cite{rotequiv:vnn}, enabling rotation to propagate through the network via simple operations. In this work, we build upon the VN framework~\cite{rotequiv:vnn} as it has been successfully applied in various rotation-aware 3D tasks~\cite{pc_regist:hegn,rotequiv:equiv_implicit}.

\subsubsection{Point Cloud Completion.}

Starting from PCN~\cite{comp:pcn}, early point cloud completion methods~\cite{comp:topnet,comp:msn,comp:crn,comp:grnet,comp:vrcnet,comp:snowflakenet} typically follow an encoder-decoder paradigm: a global feature is extracted from the partial input and decoded into a dense point cloud using generative modules such as FoldingNet~\cite{comp:folding}. This scheme inevitably leads to loss of fine-grained local details. To mitigate this, recent approaches~\cite{comp:pointr,comp:anchorformer,comp:adapointr} introduce anchor-based representations and utilize Transformer~\cite{transformer} to predict missing anchors. These methods improve geometric fidelity but are built upon rotation-variant backbones, and are mostly trained on aligned synthetic data, limiting their application in real-world scenarios with arbitrary object orientations. EquivPCN~\cite{comp_equiv:equiv_pcn} addresses this by adopting the VN framework to extract an equivariant global feature and designs an equivariant Folding operation to decode dense points. However, it still inherits the limitation of early global-feature-based designs. Other works like SCARP~\cite{comp_equiv:scarp} attempt to estimate the object pose to align inputs, but pose induction from partial data can introduce significant errors. ESCAPE~\cite{comp_equiv:escape} proposes a novel equivariant representation based on distance fields, but its evaluation of rotation-sensitive baselines follows a protocol based on test-time pose alignment rather than standard augmentation-based training. In contrast, we propose the REVNET, which bridges the gap by combining SO(3)-equivariance with anchor-based local reasoning, enabling robust and detail-preserving shape completion under arbitrary rotations.

\section{Methodology}
\label{sec:methodology}

\begin{figure}
    \centering
    \begin{tikzpicture}[font=\scriptsize, >={latex}]
        \node[inner sep=0pt] (in) {\includegraphics[height=44pt]{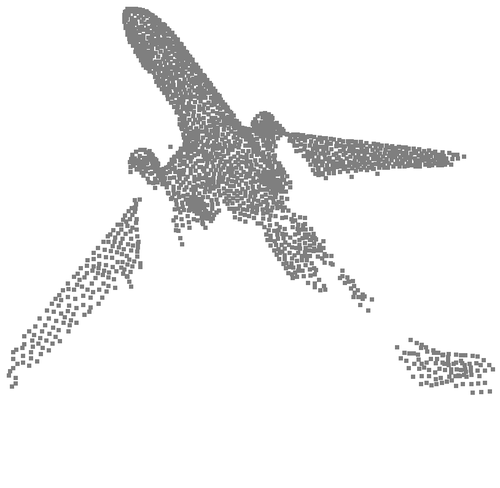}};
        \node[draw, trapezium, align=center, fill=yellow!80, very thick, anchor=north, rotate=-90, minimum height=18pt, trapezium stretches body, below right=38pt and 8pt of in] (backbone) {\rotatebox{180}{VN Backbone}};
        \node[draw, rectangle, align=center, fill=cyan!30, very thick, minimum height=50pt, minimum width=25pt, below right=22.5pt and 65pt of in, yshift=0pt] (predictor) {\parbox[c]{45pt}{\centering Missing\\Anchor\\Position\\Predictor}};
        \node[draw, rectangle, align=center, fill=cyan!80, very thick, minimum height=50pt, minimum width=25pt, right=15pt of predictor, yshift=0pt] (transformer) {\parbox[c]{45pt}{\centering VN Missing\\Anchor\\Transformer}};
        \node[draw, trapezium, align=center, fill=pink!60, very thick, anchor=north, rotate=90, minimum height=25pt, trapezium stretches body, right=60pt of transformer, xshift=-24.25pt] (decoder) {\parbox[c]{30pt}{\centering Fine\\Decoder}};
        \node[inner sep=0pt, right=245pt of in] (fine) {\includegraphics[height=44pt]{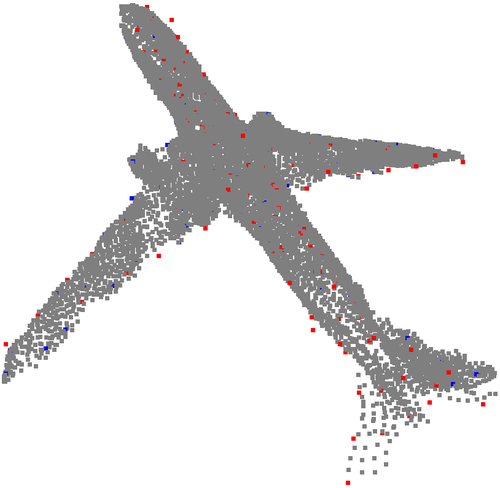}};

        \node[inner sep=0pt, right=25pt of in] (anchor) {\includegraphics[height=44pt]{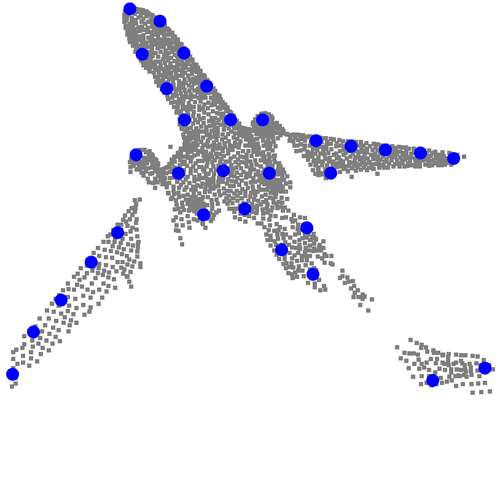}};
        \node[inner sep=0pt, left=28pt of fine] (missing) {\includegraphics[height=44pt]{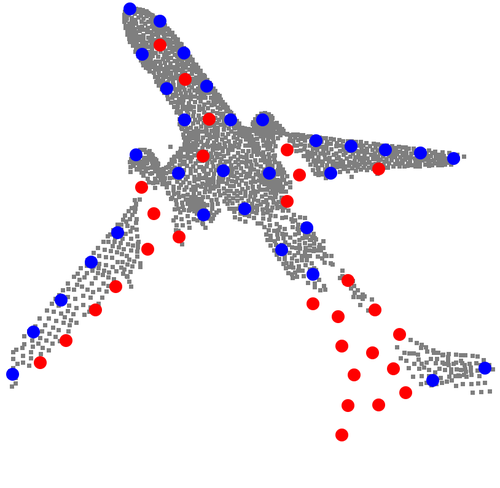}};
        
        \node[draw=none, rectangle, align=center, above=0pt of in] {Observation};
        \node[draw=none, rectangle, align=center, above=0pt of anchor] {Observed Anchors};
        \node[draw=none, rectangle, align=center, above=0pt of missing] {Predicted Anchors};
        \node[draw=none, rectangle, align=center, above=0pt of fine] {Dense Prediction};

        \node[draw=none, rectangle, align=center, below=-12pt of in] {$\mathbf{P}_{\text{in}}$};
        \node[draw=none, rectangle, inner ysep=0pt, align=center, below=-5pt of anchor] (pa){\textcolor{blue}{$\mathbf{P}_A = \{\mathbf{p}_{A,i}\}_{i=1}^N$}};
        \node[draw=none, rectangle, inner ysep=0pt, align=center, below=3pt of pa] {\textcolor{blue}{$\mathcal{X}_A = \{\mathbf{X}_{A,i}\}_{i=1}^N$}};
        \node[draw=none, rectangle, inner ysep=0pt, align=center, below=-5pt of missing] (pahat) {\textcolor{red}{$\hat{\mathbf{P}}_A = \{\hat{\mathbf{p}}_{A,j}\}_{j=1}^M$}};
        \node[draw=none, rectangle, inner ysep=0, align=center, below=1pt of pahat] {\textcolor{red}{$\hat{\mathcal{X}}_A = \{\hat{\mathbf{X}}_{A,j}\}_{j=1}^M$}};
        \node[draw=none, rectangle, align=center, below=-14pt of fine] {$\hat{\mathbf{P}}_{\text{fine}}$};

        \draw[->] ([yshift=0pt]in.south) |- ([yshift=0pt]backbone.south);
        \draw[->] ([yshift=0pt]backbone.north) -| ([yshift=-20pt, xshift=-10pt]anchor.south);
        \draw[->] ([yshift=-20pt, xshift=0pt]anchor.south) |- (predictor.west);
        \draw[->] ([xshift=0pt]predictor.east) -> ([yshift=0pt]transformer.west);
        \draw[->] ([yshift=-30pt, xshift=8pt]anchor.east) -| ([xshift=0pt]transformer.north);
        \draw[->] ([yshift=0pt]transformer.east) -| ([yshift=-20pt, xshift=5pt]missing.south);
        \draw[->] ([yshift=-20pt, xshift=15pt]missing.south) |- ([yshift=0pt]decoder.north);
        \draw[->] ([yshift=0pt]decoder.south) -| ([yshift=0pt]fine.south);

    \end{tikzpicture}
    \caption{The overview of our REVNET framework.}
    \label{fig:overview}
\end{figure}
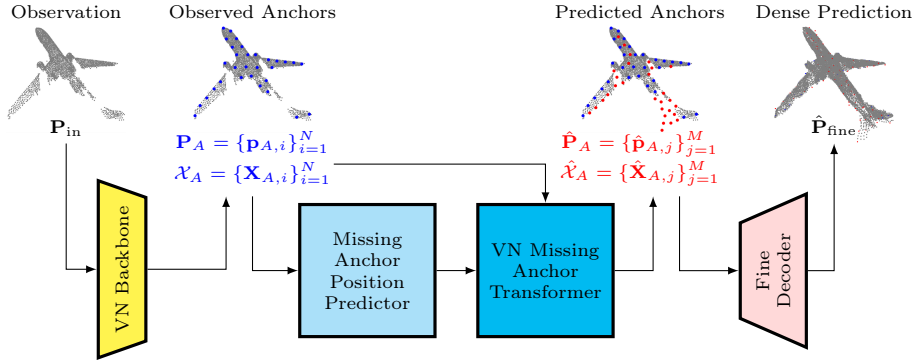

Fig.~\ref{fig:overview} provides an overview of the proposed REVNET framework. First, a novel VN-based feature backbone extracts rotation-equivariant hierarchical anchor features from the observed point cloud. Next, a missing anchor predictor estimates the position of missing anchors. A novel VN Missing Anchor Transformer (VN-MATr) then aggregates information from the existing anchors to predict the VN features for the missing anchors. Finally, a fine decoder generates dense point patches around each anchor based on a rotation-invariant Multilayer Perceptron (MLP). The details of each module are introduced in the following sections. 

\subsection{Vector Neuron Network}
We begin by briefly reviewing the Vector Neuron Network~\cite{rotequiv:vnn}, which represents a latent feature as an ordered list of 3D vectors $\mathbf{X}\in\mathbb{R}^{C\times 3}$, rather than conventional scalar features $\mathbf{x}\in\mathbb{R}^{C}$, where $C$ denotes the number of channels. This representation allows the rotation of an input point cloud to propagate naturally through network layers via a set of VN operations. As an example, the VN-Linear layer maps a VN feature from $C$ channels to $C^\prime$ channels using a learnable weight matrix $\mathbf{W}\in\mathbb{R}^{C^\prime\times C}$:
\begin{equation}
    \text{VN-Linear}(\mathbf{X}; \mathbf{W})=\mathbf{W}\mathbf{X}\text{.}
    \label{eq:vnlin}
\end{equation}
The original VN network omits the bias term, as the conventional formulation would violate rotation equivariance. In addition to VN-Linear,~\cite{rotequiv:vnn} designs a number of core VN layers, including VN-ReLU, VN-MaxPooling, and VN-BatchNorm, which can be composed into higher-level modules such as VN-MLP and VN-EdgeConv for building deeper architectures. 

An equivariant feature $\mathbf{X}\in\mathbb{R}^{C\times 3}$ can be converted to a rotation-invariant form via a VN-Inv layer. This layer first maps a $C$-channel VN feature to a $3$-channel transformation matrix $\mathbf{T} \in \mathbb{R}^{3 \times 3}$ through a VN-MLP block. The rotation-invariant feature is then obtained as $\mathbf{X}_\text{inv} = \mathbf{X}\mathbf{T}^\top$, and can later be mapped back to an equivariant form via $\mathbf{X}^\prime = \mathbf{X}_\text{inv} (\mathbf{T}^\top)^{-1}$. We refer the reader to the original paper~\cite{rotequiv:vnn} for further details.

\subsubsection{Rotation-Equivariant Bias Formulation.}
\label{sec:bias}
In conventional neural networks, a bias term shifts activation thresholds independently of the input, enhancing representational flexibility. For point cloud completion, especially in real-world scenarios, bias is beneficial when the input contains many zeros, which is common in sparse point clouds where zero-padding is used to meet a fixed input size. Consequently, grouping operations feed numerous zero vectors into the VN layers, substantially limiting expressiveness. To address this issue, we propose a novel rotation-equivariant bias formulation that preserves equivariance while restoring the representational flexibility. As shown in Eq.~\ref{eq:bias}, multiplying the mapping matrix $\mathbf{W}_B$ with $\mathbf{X}$ yields a $3 \times 3$ equivariant transformation matrix that allows the bias matrix $\mathbf{B}$ to follow the rotation of the input. To ensure stability, this transformation matrix is normalized by its Frobenius norm, preventing zero-dominant inputs from diminishing the effect of bias.
\begin{equation}
    \text{VN-Linear}_B(\mathbf{X}; \mathbf{W}, \mathbf{W}_B, \mathbf{B})=\mathbf{W}\mathbf{X} + \mathbf{B}\frac{\mathbf{W}_B\mathbf{X}}{||\mathbf{W}_B\mathbf{X}||_F}\text{.}
    \label{eq:bias}
\end{equation}

\subsubsection{ZCA-based Layer Normalization.}
\label{sec:zca}
We observe that existing normalization techniques for VN networks, such as VN-BatchNorm~\cite{rotequiv:vnn} and VN-LayerNorm~\cite{rotequiv:vntransformer}, solely normalize the norm of the vectors, which may lead to suboptimal performance due to the lack of decorrelation between the three vector dimensions. To address this limitation, we propose VN-ZCALayerNorm, a novel layer normalization scheme based on zero-phase component analysis (ZCA) that whitens the VN feature list while preserving rotation equivariance. 
Specifically, given a list of VN features $\mathcal{X}=\left[\mathbf{X}_1, \mathbf{X}_2, \dots, \mathbf{X}_N\right] \in \mathbb{R}^{N \times C \times 3}$, we first compute the mean $\mu \in \mathbb{R}^{3}$ and the covariance matrix $\Sigma \in \mathbb{R}^{3 \times 3}$ as in Eq.~\ref{eq:mu_cov}.
\begin{equation}
    \begin{split}
        & \mu = \frac{1}{NC}\sum_{i=1}^{N}\sum_{j=1}^{C}\mathbf{X}_{i}\left[j\right]\text{,} \\
        & \Sigma = \frac{1}{NC}(\mathcal{X}_f - \mu)^\top(\mathcal{X}_f - \mu)\text{,}
    \end{split}
    \label{eq:mu_cov}
\end{equation}
where $\mathbf{X}_{i}\left[j\right]$ denotes the $j$-th vector channel of the $i$-th VN feature and $\mathcal{X}_f \in \mathbb{R}^{NC \times 3}$ is the ``flattened'' VN feature list. The whitening transformation matrix is given by:
\begin{equation}
    \mathbf{W}_{\text{ZCA}} = \mathbf{U}\Lambda^{-1/2}\mathbf{U}^\top\text{,}
    \label{eq:wzca}
\end{equation}
where the matrices $\mathbf{U}$ and $\Lambda$ are originated from the eigenvalue decomposition $\Sigma=\mathbf{U}\Lambda\mathbf{U}^\top$. The formulation of the proposed VN-ZCALayerNorm is: 
\begin{equation}
    \mathbf{X}_i^\prime = (\mathbf{X}_i - \mu)\mathbf{W}_{\text{ZCA}} \odot \alpha\text{,}
    \label{eq:zcaln}
\end{equation}
where $\alpha \in \mathbb{R}^C$ is the learned scale parameter as in the standard layer normalization~\cite{layernorm}, and $\odot$ denotes channel-wise multiplication. This transformation does not affect rotation equivariance. 

\subsection{VN Feature Backbone}

Extracting distinctive and representative information from the observation is critical for point cloud completion. While deep and scalable networks have been extensively explored for other point cloud tasks such as semantic segmentation, recent point cloud completion frameworks (e.g.,~\cite{comp:pcn,comp:vrcnet,comp:pointr,comp:anchorformer,comp_equiv:equiv_pcn}) still rely on simple feature backbones such as a light-weight DGCNN. To address this gap, we design an efficient VN feature extractor to effectively capture rotation-equivariant anchor features from partial point clouds. 

\begin{figure}
    \centering
    \small
    \begin{tikzpicture}[font=\scriptsize, >={latex}]
        \node[draw=none, inner xsep=1pt, rectangle, align=center] (in) {$\mathbf{P}_{\text{in}}$};
        \node[draw, rectangle, align=center, fill=blue!20, very thick, below=8pt of in] (in_emb) {\parbox[c]{40pt}{\centering VN Input\\Embedding}};
        \node[draw, rectangle, align=center, fill=green!20, very thick, below=13pt of in_emb] (sa) {\parbox[c]{35pt}{\centering VN-\\SA}};
        \node[draw, rectangle, align=center, fill=yellow!50, very thick, below=10pt of sa] (resmlp) {\parbox[c]{35pt}{\centering VN-\\ResMLP}};
        \node[draw=none, rectangle, inner ysep=0pt, align=center, below=13pt of resmlp] (out) {$\mathbf{P}_A$, $\mathcal{X}_A$};
        \node[draw=none, rectangle, align=center, left=-1pt of sa] {$\times 1$};
        \node[draw=none, rectangle, align=center, left=-1pt of resmlp] {$\times$any};
        \node[draw, color=gray, rectangle, align=center, fill=none, dashed, very thick, minimum height=63pt, minimum width=68pt, below=5pt of in_emb, xshift=-9pt] (block_box) {};
        \node[draw=none, rectangle, align=center, fill=none, above right=-10pt and -1pt of block_box] {$\times 3$};
        \draw[->] (in.south) -> (in_emb.north);
        \draw[->] (in_emb.south) -> (sa.north);
        \draw[->] (sa.south) -> (resmlp.north);
        \draw[->] (resmlp.south) -> (out.north);
        \node[draw=none, rectangle, align=center, fill=none, above=-2pt of in, xshift=-5pt] (backbone_title) {\textbf{VN Feature Backbone}};
        \node[draw=none, fill=green!20, rectangle, align=center, minimum height=140pt, minimum width=105pt, below right=-23pt and 45pt of in] (sa_box) {};
        \node[draw=none, rectangle, align=center, fill=none, right=43pt of backbone_title] (sa_title) {\textbf{VN-SA}};
        \node[draw=none, inner ysep=1pt, rectangle, align=center, right=75pt of in] (sa_in) {$\mathcal{X}$};
        \node[draw, rectangle, align=center, fill=blue!20, very thick, below=7pt of sa_in] (sa_mlp_in) {\parbox[c]{50pt}{\centering VN-MLP}};
        \node[draw, rectangle, align=center, fill=blue!20, very thick, below=7pt of sa_mlp_in] (sa_down) {\parbox[c]{50pt}{\centering FPS\\Downsample}};
        \node[draw, rectangle, align=center, fill=blue!20, very thick, below=7pt of sa_down] (sa_group) {\parbox[c]{50pt}{\centering Grouping}};
        \node[draw, circle, align=center, fill=none, inner sep=0, thick, minimum height=3pt, minimum width=3pt, below=5pt of sa_group] (sa_plus) {$+$};
        \node[draw, rectangle, align=center, fill=blue!20, very thick, below=5pt of sa_plus] (sa_pool) {\parbox[c]{50pt}{\centering VN-Pooling}};
        \node[draw=none, inner ysep=0pt, rectangle, align=center, below=7pt of sa_pool] (sa_out) {$\mathcal{X}^\prime_{\text{down}}$, $\mathbf{P}_{\text{down}}$};
        \node[draw=none, inner ysep=1pt, rectangle, align=center, right=35pt of sa_in] (sa_in_p) {$\mathbf{P}$};
        \node[draw, rectangle, align=center, fill=blue!20, very thick, below=55.25pt of sa_in_p] (sa_pos) {\parbox[c]{18pt}{\centering RPE}};
        \draw[->] (sa_in.south) -> (sa_mlp_in.north);
        \draw[->] (sa_mlp_in.south) -> (sa_down.north);
        \draw[->] (sa_down.south) -> (sa_group.north);
        \draw[->] (sa_group.south) -> (sa_plus.north);
        \draw[->] (sa_plus.south) -> (sa_pool.north);
        \draw[->] (sa_pool.south) -> (sa_out.north);
        \draw[->] (sa_group.east) -> (sa_pos.west);
        \draw[->] (sa_in_p.south) -> (sa_pos.north);
        \draw[->] (sa_pos.south) |- (sa_plus.east);
        \node[draw=none, fill=yellow!50, rectangle, align=center, minimum height=140pt, minimum width=115pt, right=0pt of sa_box] (mlp_box) {};
        \node[draw=none, rectangle, align=center, fill=none, right=62pt of sa_title] {\textbf{VN-ResMLP}};
        \node[draw=none, inner ysep=1pt, rectangle, align=center, right=100pt of sa_in] (mlp_in) {$\mathcal{X}$};
        \node[draw, rectangle, align=center, fill=blue!20, very thick, below=7pt of mlp_in] (mlp_mlp_in) {\parbox[c]{50pt}{\centering VN-MLP}};
        \node[draw, rectangle, align=center, fill=blue!20, very thick, below=7pt of mlp_mlp_in] (mlp_group) {\parbox[c]{50pt}{\centering Grouping}};
        \node[draw, circle, align=center, fill=none, inner sep=0, thick, minimum height=3pt, minimum width=3pt, below=5pt of mlp_group] (mlp_plus) {$+$};
        \node[draw, rectangle, align=center, fill=blue!20, very thick, below=5pt of mlp_plus] (mlp_pool) {\parbox[c]{50pt}{\centering VN-Pooling}};
        \node[draw, rectangle, align=center, fill=blue!20, very thick, below=5pt of mlp_pool] (mlp_mlp_update) {\parbox[c]{50pt}{\centering VN-MLP}};
        \node[draw, circle, align=center, fill=none, inner sep=0, thick, minimum height=3pt, minimum width=3pt, below=5pt of mlp_mlp_update] (mlp_res) {$+$};
        \node[draw=none, inner ysep=0pt, rectangle, align=center, below=7pt of mlp_res] (mlp_out) {$\mathcal{X}^\prime$};
        \node[draw=none, inner ysep=1pt, rectangle, align=center, right=35pt of mlp_in] (mlp_in_p) {$\mathbf{P}$};
        \node[draw, rectangle, align=center, fill=blue!20, very thick, below=26.75pt of mlp_in_p] (mlp_pos) {\parbox[c]{18pt}{\centering RPE}};
        \draw[->] (mlp_in.south) -> (mlp_mlp_in.north);
        \draw[->] (mlp_mlp_in.south) -> (mlp_group.north);
        \draw[->] (mlp_group.south) -> (mlp_plus.north);
        \draw[->] (mlp_plus.south) -> (mlp_pool.north);
        \draw[->] (mlp_pool.south) -> (mlp_mlp_update.north);
        \draw[->] (mlp_mlp_update.south) -> (mlp_res.north);
        \draw[->] (mlp_res.south) -> (mlp_out.north);
        \draw[->] (mlp_group.east) -> (mlp_pos.west);
        \draw[->] (mlp_in_p.south) -> (mlp_pos.north);
        \draw[->] (mlp_pos.south) |- (mlp_plus.east);
        \draw[->] (mlp_in.west) --++ (-28pt, 0) |- (mlp_res.west);
        
    \end{tikzpicture}
    \caption{The architecture of our VN feature backbone. }
    \label{fig:backbone}
\end{figure}
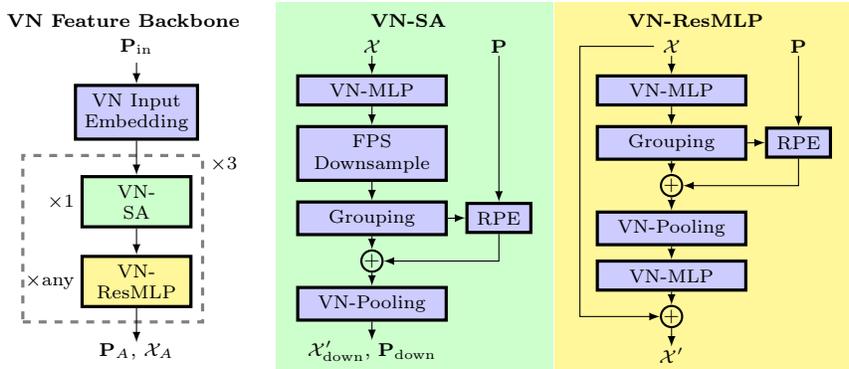

As illustrated in Fig.~\ref{fig:backbone}, we adopt a multi-stage scheme to hierarchically downsample the input point cloud $\mathbf{P}_\text{in}\in\mathbb{R}^{N_{in} \times 3}$ to a set of anchor positions $\mathbf{P}_A = \{\mathbf{p}_{A,i}\}_{i=1}^N$ and extract corresponding equivariant anchor features $\mathcal{X}_A = \{\mathbf{X}_{A,i}\}_{i=1}^N$, where $N$ denotes the number of observed anchors. Directly applying a VN-Linear layer to a 3D coordinate $\mathbf{p}\in \mathbb{R}^{1\times 3}$ results in vector neurons that are all linearly dependent~\cite{rotequiv:vnn}. To mitigate this, we apply a VN-EdgeConv layer using $k_0$ neighbors to lift each input point to a $C_0$-channel VN input embedding $\mathbf{X}_{\text{in}, i} = \text{VN-EdgeConv}(\mathbf{p}_{\text{in}, i}; k_0)$, which serves as input to subsequent stages. Each stage in our backbone consists of one set-abstraction block (VN-SA), followed by several residual VN-MLP blocks (VN-ResMLP). The VN-SA block downsamples a point cloud using the farthest point sampling algorithm and aggregate features in the local neighborhood, while the VN-ResMLP enhances the local features. We follow the best practice discovered in~\cite{3d_point:pointmetabase} to perform neighbor updates before grouping and incorporate a relative position encoding (RPE) to preserve local geometric relationships: $\text{RPE}(\mathbf{p}_i, \mathbf{p}_j)=\text{VN-MLP}(\mathbf{p}_i - \mathbf{p}_j)$. Finally, for each anchor $A_i$, features from all stages are concatenated and fused via a VN-MLP to produce the final anchor representation. 

\subsection{Missing Anchor Position Predictor}
Based on the observed anchors, the next step is to predict the positions and features for the missing anchors. As noted in~\cite{rotequiv:vnn}, directly using a VN-MLP to generate coarse point clouds, as commonly done in non-equivariant models, can lead to instability. This is because mapping a VN feature to a 3D coordinate, i.e., $\mathbb{R}^{C \times 3} \rightarrow \mathbb{R}^{1 \times 3}$, shares the same weight matrix across all vector dimensions to maintain rotation equivariance. We also observe that the DPK module proposed in~\cite{rotequiv:vnn} introduces significant noise during generation of anchor positions.
 
Therefore, we leverage the flexible equivariance-invariance conversion property of VN features to generate missing anchor positions $\hat{\mathbf{P}}_A = \{\hat{\mathbf{p}}_{A,j}\}_{j=1}^M$ in a more stable manner. First, we lift the observed anchor features to a higher-dimensional VN space and aggregate them into a global feature $\mathbf{X}_g \in \mathbb{R}^{C_g \times 3}$ via a pooling layer. This global feature is then passed through a VN-Inv layer to obtain a rotation-invariant representation $\mathbf{X}_{g,\text{inv}}$. We then predict the 3D coordinates of the $M$ missing anchors in a canonical, rotation-invariant frame using a conventional MLP. Finally, we transform these coordinates back to the original frame using the inverse matrix derived during the VN-Inv step, thereby restoring their equivariance. This approach improves stability and preserves geometric consistency under arbitrary rotations.

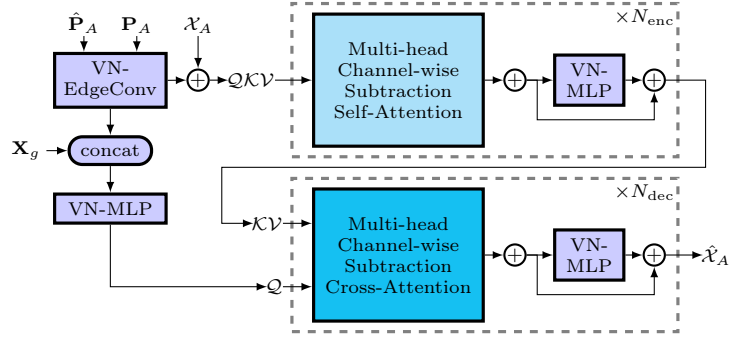
\begin{figure}
    \centering
    \small
    \begin{tikzpicture}[font=\scriptsize, >={latex}]
        \node[draw=none, inner ysep=1pt, rectangle, align=center] (in_pred) {$\hat{\mathbf{P}}_{A}$};
        \node[draw, rectangle, align=center, fill=blue!20, very thick, below=6pt of in_pred, xshift=10pt] (in_emb) {\parbox[c]{36pt}{\centering VN-\\EdgeConv}};
        \node[draw=none, inner ysep=1pt, rectangle, align=center, above=6pt of in_emb, xshift=10pt] (in_obs) {$\mathbf{P}_{A}$};
        \node[draw, rectangle, align=center, rounded corners=5pt, fill=blue!20, very thick, below=10pt of in_emb] (cat) {\parbox[c]{25pt}{\centering concat}};
        \node[draw=none, inner ysep=1pt, rectangle, align=center, left=7pt of cat] (xg) {$\mathbf{X}_g$};
        \node[draw, rectangle, align=center, fill=blue!20, very thick, below=10pt of cat] (q_mlp) {\parbox[c]{36pt}{\centering VN-MLP}};
        \node[draw=none, inner ysep=1pt, rectangle, align=center, right=5pt of in_obs] (in_x) {$\mathcal{X}_{A}$};
        \node[draw, circle, align=center, fill=none, inner sep=0, thick, minimum height=3pt, minimum width=3pt, below=12.3pt of in_x] (emb_plus) {$+$};
        \node[draw=none, inner xsep=1pt, rectangle, align=center, right=5pt of emb_plus] (qkv) {$\mathcal{QKV}$};
        \node[draw, rectangle, align=center, fill=cyan!30, very thick, minimum height=50pt, right=13pt of qkv] (sa) {\parbox[c]{58pt}{\centering Multi-head\\Channel-wise\\Subtraction\\Self-Attention}};
        \node[draw, circle, align=center, fill=none, inner sep=0, thick, minimum height=3pt, minimum width=3pt, right=7pt of sa] (sa_res1) {$+$};
        \node[draw, rectangle, align=center, fill=blue!20, very thick, right=10pt of sa_res1] (sa_mlp) {\parbox[c]{20pt}{\centering VN-\\MLP}};
        \node[draw, circle, align=center, fill=none, inner sep=0, thick, minimum height=3pt, minimum width=3pt, right=6.5pt of sa_mlp] (sa_res2) {$+$};
        \node[draw, color=gray, rectangle, align=center, fill=none, dashed, very thick, minimum height=58pt, minimum width=146pt, left=-139pt of sa] (sa_box) {};
        \node[draw=none, rectangle, align=center, fill=none, above right=-12pt and -28pt of sa_box] {$\times N_{\text{enc}}$};
        \node[draw, rectangle, align=center, fill=cyan!70, very thick, minimum height=50pt, below=15pt of sa] (ca) {\parbox[c]{58pt}{\centering Multi-head\\Channel-wise\\Subtraction\\Cross-Attention}};
        \node[draw, circle, align=center, fill=none, inner sep=0, thick, minimum height=3pt, minimum width=3pt, right=7pt of ca] (ca_res1) {$+$};
        \node[draw, rectangle, align=center, fill=blue!20, very thick, right=10pt of ca_res1] (ca_mlp) {\parbox[c]{20pt}{\centering VN-\\MLP}};
        \node[draw, circle, align=center, fill=none, inner sep=0, thick, minimum height=3pt, minimum width=3pt, right=6.5pt of ca_mlp] (ca_res2) {$+$};
        \node[draw=none, inner xsep=1pt, rectangle, align=center, right=12pt of ca_res2] (out) {$\hat{\mathcal{X}}_{A}$};
        \node[draw, color=gray, rectangle, align=center, fill=none, dashed, very thick, minimum height=58pt, minimum width=146pt, left=-139pt of ca] (ca_box) {};
        \node[draw=none, rectangle, align=center, fill=none, above right=-12pt and -28pt of ca_box] {$\times N_{\text{dec}}$};

        \node[draw=none, inner xsep=1pt, rectangle, align=center, left=10pt of ca, yshift=12pt] (kv) {$\mathcal{KV}$};
        \node[draw=none, inner xsep=1pt, rectangle, align=center, left=10pt of ca, yshift=-12pt] (q) {$\mathcal{Q}$};

        \draw[->] (in_pred.south) -> ([xshift=-10pt]in_emb.north);
        \draw[->] (in_obs.south) -> ([xshift=10pt]in_emb.north);
        \draw[->] (in_emb.south) -> (cat.north);
        \draw[->] (cat.south) -> (q_mlp.north);
        \draw[->] ([xshift=-1pt]xg.east) -> (cat.west);
        \draw[->] (in_x.south) -> (emb_plus.north);
        \draw[->] (in_emb.east) -> (emb_plus.west);
        \draw[->] (emb_plus.east) -> ([xshift=2pt]qkv.west);

        \draw[->] ([xshift=-1pt]qkv.east) -> (sa.west);
        \draw[->] (sa.east) -> (sa_res1.west);
        \draw[->] (sa_res1.east) -> (sa_mlp.west);
        \draw[->] (sa_res1.east) --++ (4pt, 0) --++ (0, -15pt) -| (sa_res2.south);
        \draw[->] (sa_mlp.east) -> (sa_res2.west);

        \draw[->] (sa_res2.east) --++ (15pt, 0) --++ (0, -33pt) --++ (-183pt, 0) |- ([xshift=2pt]kv.west);

        \draw[->] (q_mlp.south) |- ([xshift=2pt]q.west);
        \draw[->] ([xshift=-1pt]q.east) -> ([yshift=-12pt]ca.west);
        \draw[->] ([xshift=-1pt]kv.east) -> ([yshift=12pt]ca.west);
        \draw[->] (ca.east) -> (ca_res1.west);
        \draw[->] (ca_res1.east) -> (ca_mlp.west);
        \draw[->] (ca_res1.east) --++ (4pt, 0) --++ (0, -15pt) -| (ca_res2.south);
        \draw[->] (ca_mlp.east) -> (ca_res2.west);
        \draw[->] (ca_res2.east) -> ([xshift=2pt]out.west);

    \end{tikzpicture}
    \caption{The architecture of VN Missing Anchor Transformer. }
    \label{fig:vnmatr}
\end{figure}

\subsection{VN Missing Anchor Transformer}
We design a VN-based Missing Anchor Transformer (VN-MATr) to infer equivariant features for missing anchors by using contextual information from the observed anchors. Fig.~\ref{fig:vnmatr} illustrates the architecture of VN-MATr. For each predicted anchor position $\hat{\mathbf{p}}_{A,j}$, similar to the input embedding in the feature backbone, we first lift 1-channel coordinates to a high-dimensional VN feature using their $k_A$ nearest neighbor anchors: $\hat{\mathbf{X}}_{\text{emb},j} = \text{VN-EdgeConv}(\hat{\mathbf{p}}_{A,j}; k_A)$. This positional embedding is then concatenated with the global feature $\mathbf{X}_g$ and a shared VN-MLP is applied to produce the query embedding: 
\begin{equation}
    \mathbf{Q}_j = \text{VN-MLP}(\text{concat}[\mathbf{X}_g, \hat{\mathbf{X}}_{\text{emb},j}])\text{.}
    \label{eq:q}
\end{equation}

Our VN-MATr adopts an encoder-decoder Transformer architecture~\cite{transformer}. The encoder with $N_\text{enc}$ blocks employs the self-attention to enhance the observed anchor features $\mathcal{X}_{A}$. The output of the encoder is served as the keys and values $\mathcal{X}^\prime_{A}=\mathcal{K} = \mathcal{V} = \{\mathbf{X^\prime}_{A,i}\}_{i=1}^N \in \mathbb{R}^{N \times C \times 3}$ to the decoder, while the query embeddings $\mathcal{Q} = \{\mathbf{Q}_j\}_{j=1}^M \in \mathbb{R}^{M \times C \times 3}$ of missing anchors are served as the queries. Through $N_{\text{dec}}$ decoder layers, the query embeddings are transformed into the final predicted features for the missing anchors, denoted as $\hat{\mathcal{X}}_A = \{\hat{\mathbf{X}}_{A,j}\}_{j=1}^M$.

Unlike Frobenius inner product attention employed in VN-Transformer~\cite{rotequiv:vntransformer}, we adopt a channel-wise subtraction attention (CWSA) similar to~\cite{3d_point:pt}. For a query $\mathbf{Q}_j$, the $N \times C$ score matrix is computed as:
\begin{equation}
    \textbf{Att}(\mathbf{Q}_j, \mathcal{K}) = \text{softmax}(\left[\text{MLP}(\text{VN-Inv}(\mathbf{Q}_j - \mathbf{K}_i))\right]_{i=1}^N)\text{,}
\end{equation}
where the MLP maps the rotation-invariant relation feature to a $C$-dimensional score vector. The weighted aggregation of observed anchor features for query $\mathbf{Q}_j$ is performed as:
\begin{equation}
    \text{VN-Sub-Attn}(\mathbf{Q}_j, \mathcal{K}, \mathcal{V}) = \sum_{i=1}^N\textbf{Att}(\mathbf{Q}_j, \mathcal{K})\left[i\right] \odot \mathbf{V}_i\text{.}
\end{equation}
This formulation allows each vector channel to receive distinct attention weights, enhancing the expressiveness of the Transformer while preserving rotation equivariance, since the attention scores are invariant and applied per channel. Furthermore, this mechanism can be extended to a multi-head CWSA variant, where each head employs its own VN-Inv and MLP layers, enabling the model to attend to different subspaces in a rotation-equivariant manner.

\subsection{Local Rotation-Invariant Fine Decoder}
The final stage of REVNET involves decoding both the observed and predicted equivariant anchor features into a dense point cloud. Similar to the missing anchor position predictor, we first convert all equivariant anchor features into rotation-invariant representations using individual transformation matrices produced by a shared VN-Inv layer. In the resulting rotation-invariant frame, we adopt a simple coarse-to-fine MLP to generate dense point offsets around each anchor. These offsets are then transformed back to the original equivariant frames and added to the corresponding anchor positions. This yields the final predicted complete point cloud $\hat{\mathbf{P}}_{\text{fine}}$.

\subsection{Loss Function}
The weights of REVNET are optimized end-to-end using the commonly adopted L1 Chamfer Distance (CD-$l_1$)~\cite{3d_point:chamfer_distance} as the loss function. Chamfer Distance measures the similarity between two point sets and is invariant to the permutation of points. Given two point clouds $\mathbf{P}_1$ and $\mathbf{P}_2$, CD-$l_1$ is computed as:
\begin{equation}
    d_{\text{CD-}l_1}(\mathbf{P}_1, \mathbf{P}_2) = \frac{1}{\left|\mathbf{P}_1\right|}\sum_{x \in \mathbf{P}_1}\min_{y \in \mathbf{P}_2}\left\|x-y\right\|_2 + \frac{1}{\left|\mathbf{P}_2\right|}\sum_{y \in \mathbf{P}_2}\min_{x \in \mathbf{P}_1}\left\|x-y\right\|_2\text{.}
    \label{eq:cdl1}
\end{equation}
We use the ground-truth dense point cloud $\mathbf{P}_{\text{gt}}$ to supervise both the predictions of anchor positions and dense points. 

\section{Experiments}
\label{sec:exps}

\subsection{Datasets and Baselines}
We primarily train and evaluate our model on the synthetic MVP dataset~\cite{comp:vrcnet}, which contains partial point clouds sampled from multiple views of 3D CAD models. Following~\cite{comp_equiv:equiv_pcn}, we use the same eight object categories with an output resolution of $8192$ points. The performance is evaluated using CD-$l_1$ and F-Score at both strict tolerance of $0.01$ (F-Score@1\%) and relaxed tolerance of $0.02$ (F-Score@2\%). Additionally, we measure the completion consistency under rotations following~\cite{comp_equiv:equiv_pcn}. For each model, the partial-complete point cloud pair is randomly rotated for $30$ times, and the score $cst_{\text{CD-}l_1}$ is defined as the largest CD-$l_1$ difference across these evaluations.

We compare our REVNET with both rotation-equivariant and rotation-variant baselines on the MVP dataset. Rotation-variant models are trained with rotation augmentation using random angles between $\pm 180^{\circ}$ along all three axes (i.e., SO(3)/SO(3)), while rotation-equivariant methods are trained without augmentation (i.e., None/SO(3)). All evaluations are conducted on inputs under random SO(3) rotations. Results for earlier methods~\cite{comp:pcn,comp:topnet,comp:msn,comp:crn,comp:grnet,comp:vrcnet,comp:snowflakenet,comp_equiv:equiv_pcn} are taken from~\cite{comp_equiv:equiv_pcn}. The F-score@1\% for these models was not provided. More recent rotation-variant methods~\cite{comp:anchorformer,comp:pointr,comp:gtnet,comp:pmpnet++,comp:odgnet,comp:adapointr} are re-trained using their released network architectures and training configurations under the SO(3)/SO(3) augmentation setup. The rotation-equivariant model ESCAPE~\cite{comp_equiv:escape} is re-trained without rotation augmentation.

To assess our model's generalization to real-world scenarios, we test it on the KITTI dataset~\cite{data:kitti}, which contains sparse outdoor LiDAR scans of cars. We follow the standard practice from~\cite{comp:pointr,comp:anchorformer} to train our model on the synthetic PCN Car dataset~\cite{comp:pcn} and test it on all KITTI partial shapes, using an output resolution of 16384 points. To support rotation-variant models, the KITTI partial point clouds are normalized and rotated into a canonical pose using hand-labeled bounding boxes (i.e., None/None). The results are reported in terms of two metrics: Fidelity Distance (FD), defined as the one-directional CD-$l_2$ from the partial observation to the predicted completion; and Minimal Matching Distance (MMD), which measures the minimal CD-$l_2$ between the predicted point cloud and all car models from a synthetic reference dataset. 

The baseline results of rotation-variant models on the KITTI dataset are taken from~\cite{comp:pointr,comp:anchorformer}. We re-train ESCAPE using the hyperparameters provided by the authors under the same benchmark setup as our model, obtaining results that are comparable to, or better than, those reported in the original paper. EquivPCN is excluded from this comparison because no pretrained weights are publicly available, and our attempts to re-train the model under settings similar to those used for synthetic datasets did not converge.

\subsection{Implementation Details}
Our framework is implemented using PyTorch. The feature backbone extracts $N=128$ anchors from the input partial point cloud. The VN-MATr module with 4 encoder layers and 6 decoder layers predicts $M=128$ anchors for the missing region. For MVP, we train our model on an NVIDIA RTX 5090 GPU with a batch size of 32 for 100 epochs using AdamW optimizer. We set the initial learning rate to 0.0005 and apply a learning rate decay of 0.7 for every 20 epochs. 

\setlength{\tabcolsep}{1.5pt}
\begin{table}
    \centering
    \small
    \begin{tabular}{c|c c c c c c c c|c||c|c}
        \hline
        & \multicolumn{9}{c||}{CD-$l_1 \times 100 \downarrow$} & \multicolumn{2}{c}{F-Score $\uparrow$} \\
        \hline
        \hline
        Method & Air & Cab & Car & Cha & Lam & Sof & Tab & Ves & Avg & Avg@1\% & Avg@2\% \\
        \hline
        & \multicolumn{11}{c}{SO(3)/SO(3)} \\
        \hline
        PCN & 1.01 & 1.55 & 1.38 & 1.81 & 1.50 & 1.66 & 1.82 & 1.32 & 1.51 &
        - & 77.82 \\
        TopNet & 1.40 & 1.88 & 1.58 & 2.24 & 1.84 & 1.96 & 2.22 & 1.53 & 1.83 &
        - & 67.20 \\
        MSN & 0.90 & 1.55 & 1.34 & 1.60 & 1.16 & 1.58 & 1.50 & 1.27 & 1.36 &
        - & 79.14 \\
        CRN & 0.86 & 1.52 & 1.30 & 1.73 & 1.34 & 1.61 & 1.57 & 1.18 & 1.39 &
        - & 79.71 \\
        GRNet & 0.90 & 1.64 & 1.37 & 1.47 & 1.17 & 1.59 & 1.36 & 1.24 & 1.39 &
        - & 81.50 \\
        VRCNet & 0.65 & 1.31 & 1.13 & 1.25 & 0.94 & 1.30 & 1.17 & 1.03 & 1.10 &
        - & 86.28 \\
        SnowFlake & 0.63 & 1.25 & 1.09 & 1.16 & 0.88 & 1.20 & 1.05 & 0.97 & 1.03 &
        - & - \\
        PoinTr & 0.84 & 1.44 & 1.26 & 1.47 & 1.12 & 1.45 & 1.40 & 1.15 & 1.27 &
        56.69 & 82.58 \\
        AnchorFormer & 0.79 & 1.48 & 1.49 & 1.50 & 1.20 & 1.73 & 1.25 & 1.50 & 1.37 &
        55.83 & 82.44 \\
        PMP-Net++ & 0.62 & 1.40 & 1.21 & 1.18 & 0.85 & 1.31 & 1.09 & 0.96 & 1.08 &
        60.47 & 81.62 \\
        GTNet & 0.62 & 1.30 & 1.17 & 1.19 & 0.91 & 1.27 & 1.11 & 9.72 & 1.07 &
        61.17 & 85.53 \\
        ODGNet & 0.55 & 1.18 & 1.05 & 1.07 & 0.78 & 1.14 & 1.01 & 0.89 & 0.96 &
        67.61 & 89.52 \\
        AdaPoinTr & 0.53 & 1.15 & 1.04 & 1.05 & 0.79 & 1.13 & 0.96 & \textbf{0.85} & 0.94 &
        68.02 & 90.80 \\

        \hline
        & \multicolumn{11}{c}{None/SO(3)} \\
        \hline
        EquivPCN & 0.54 & 1.18 & 1.04 & 1.11 & 0.89 & 1.17 & 1.03 & 0.92 & 0.98 &
        67.52 & 88.96 \\
        ESCAPE & 0.72 & 1.41 & 1.19 & 1.26 & 1.08 & 1.41 & 1.20 & 1.20 & 1.18 &
        61.53 & 85.92 \\
        Ours & \textbf{0.50} & \textbf{1.08} & \textbf{0.96} & \textbf{0.98} & \textbf{0.77} & \textbf{1.05} & \textbf{0.90} & \textbf{0.85} & \textbf{0.89} &
        \textbf{73.44} & \textbf{92.37} \\
        \hline
    \end{tabular}
    \caption{Completion accuracy in terms of CD-$l_1 \times 100 \downarrow$, F-Score@1\% $\uparrow$, and F-Score@2\% $\uparrow$ on the MVP dataset with a resolution of 8192 points. The best results are highlighted in bold. Our model achieves the best performance in almost all categories. }
    \label{tab:mvp_cd}
\end{table}

\subsection{Results and Discussion}

Table~\ref{tab:mvp_cd} reports point cloud completion accuracy on the MVP dataset. Our model achieves the best performance across almost all object categories, demonstrating its effectiveness for SO(3)-equivariant point cloud completion. As shown in Table~\ref{tab:mvp_cst}, our model also attains the lowest $cst_{\text{CD-}l_1}$, indicating its ability to produce highly consistent shapes under arbitrary rotations. Moreover, with deterministic farthest point sampling, the consistency further improves to the order of $10^{-6}$, validating the strict equivariance of our framework. Interestingly, recent advanced rotation-variant methods such as ODGNet~\cite{comp:odgnet} and AdaPoinTr~\cite{comp:adapointr} already surpass the completion accuracy of equivariant baselines, showing their ability to synthesize plausible points under arbitrary rotations through extensive augmentation. However, their completion consistency remains relatively high, suggesting that such models do not fully capture the principle of equivariance.

\begin{figure}
    \centering
    \scriptsize
    \renewcommand{\arraystretch}{0.0}
    \begin{tabular}{m{30pt}<{\centering} m{30pt}<{\centering} m{30pt}<{\centering} m{30pt}<{\centering} m{30pt}<{\centering} m{30pt}<{\centering} m{30pt}<{\centering} m{30pt}<{\centering} m{30pt}<{\centering} m{30pt}<{\centering}}
        Input & Anchor-Former & PMP-Net++ & GTNet & ODG-Net & Ada-PoinTr & Equiv-PCN & ESCAPE & Ours & Ground-truth\\

        \includegraphics[width=30pt]{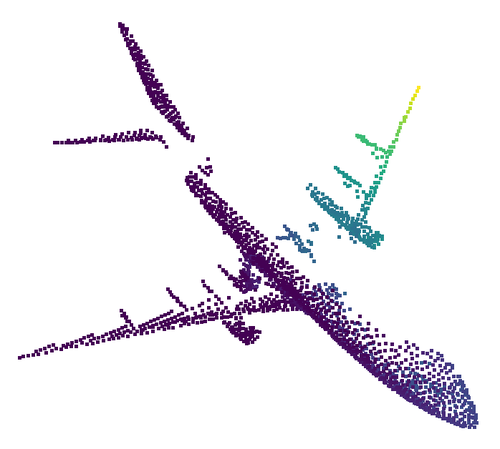} & 
        \includegraphics[width=30pt]{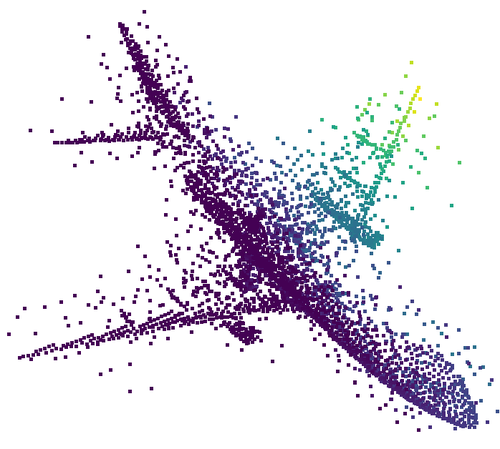} & 
        \includegraphics[width=30pt]{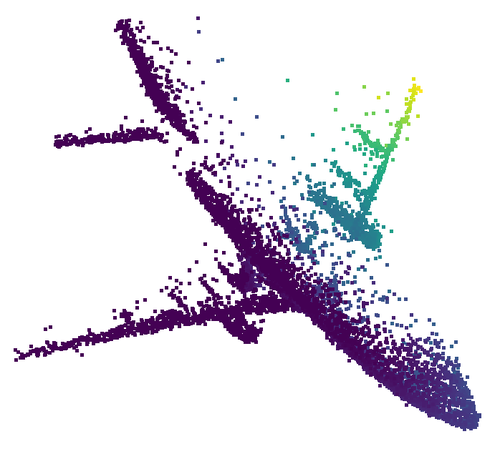} & 
        \includegraphics[width=30pt]{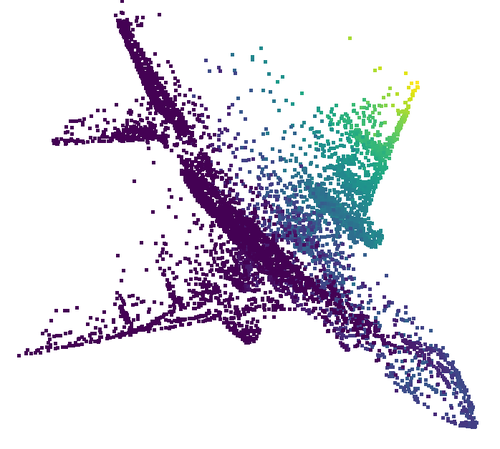} & 
        \includegraphics[width=30pt]{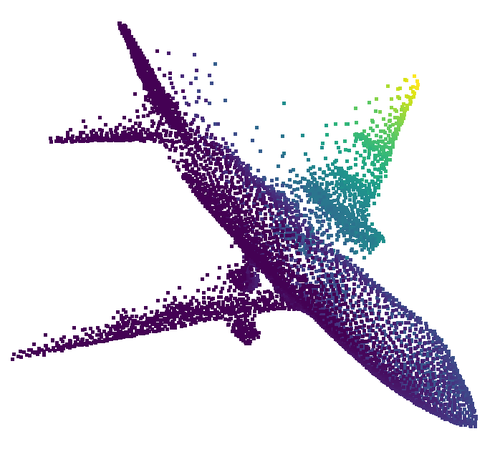} & 
        \includegraphics[width=30pt]{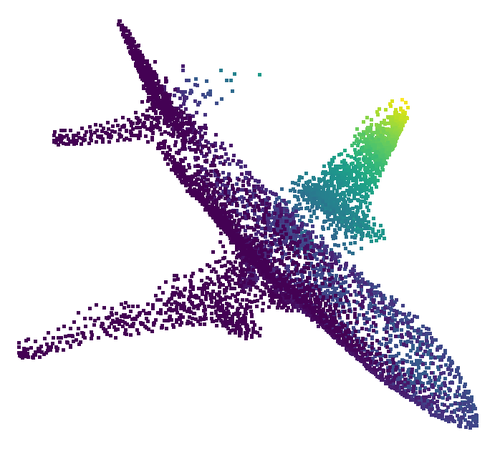} & 
        \includegraphics[width=30pt]{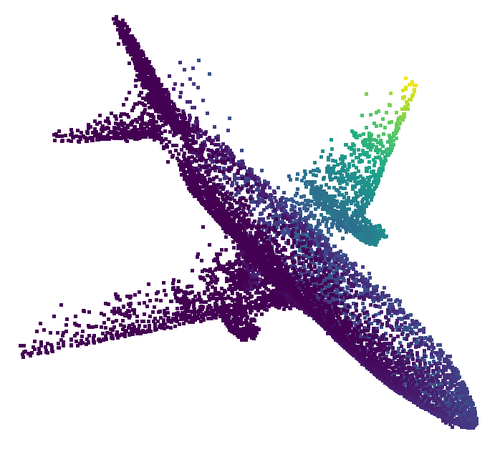} & 
        \includegraphics[width=30pt]{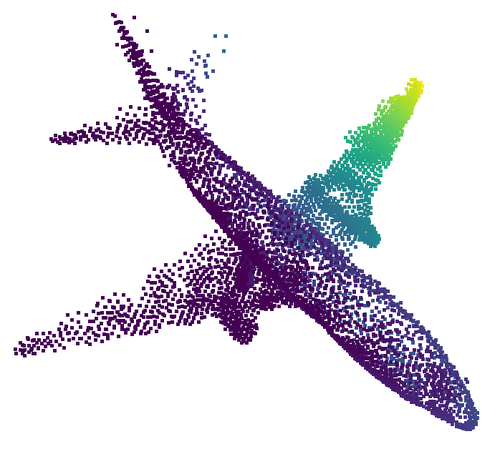} & 
        \includegraphics[width=30pt]{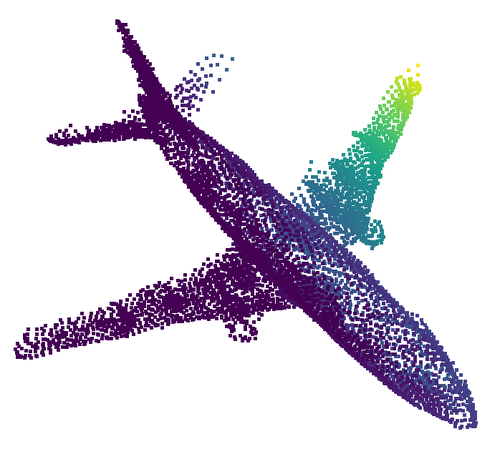} & 
        \includegraphics[width=30pt]{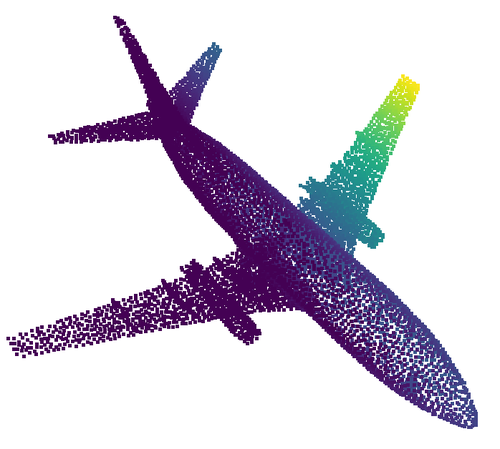} \\

        \includegraphics[width=30pt]{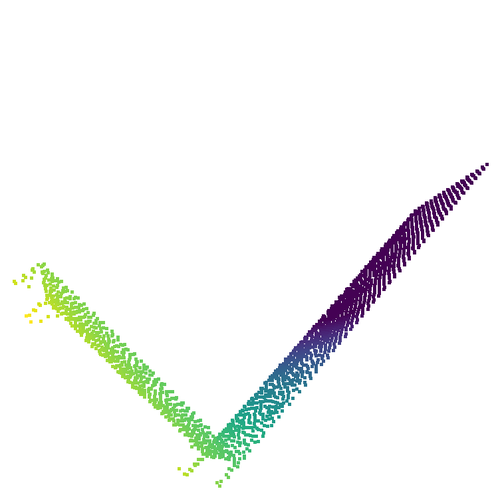} & 
        \includegraphics[width=30pt]{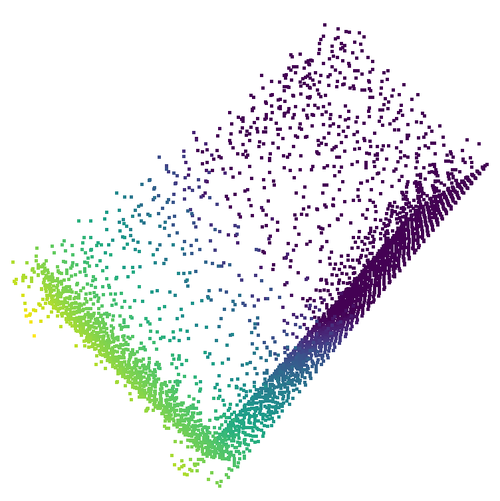} & 
        \includegraphics[width=30pt]{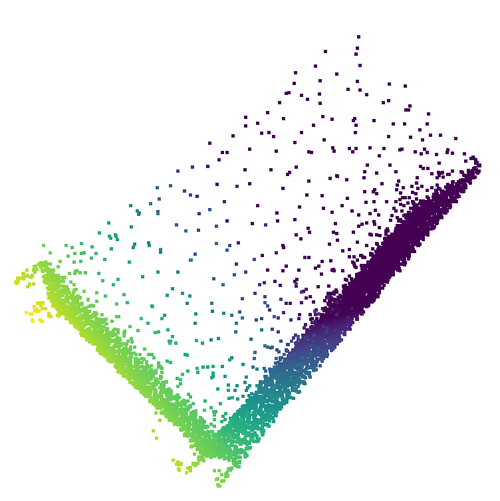} & 
        \includegraphics[width=30pt]{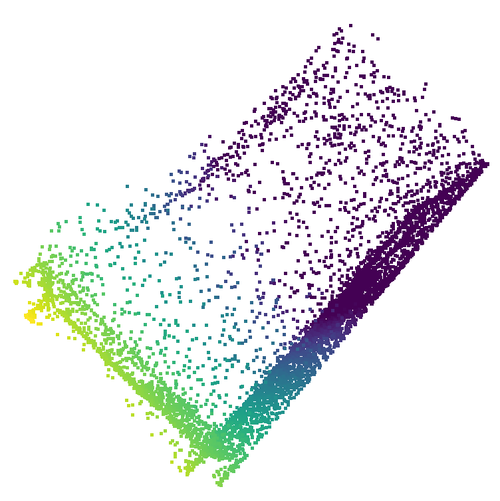} & 
        \includegraphics[width=30pt]{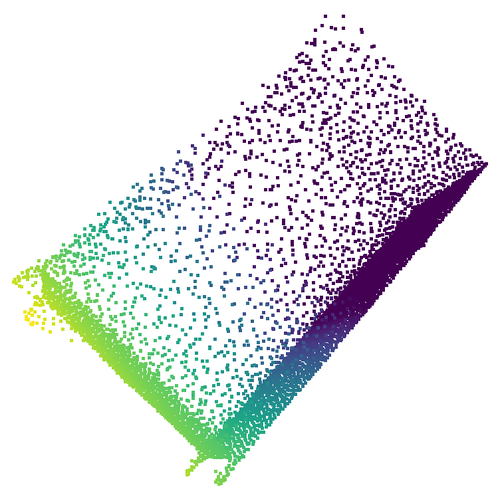} & 
        \includegraphics[width=30pt]{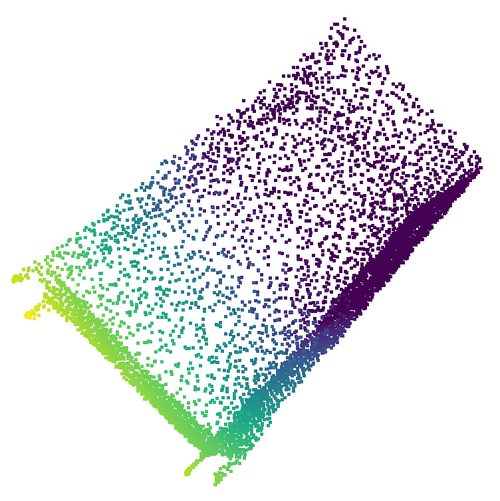} & 
        \includegraphics[width=30pt]{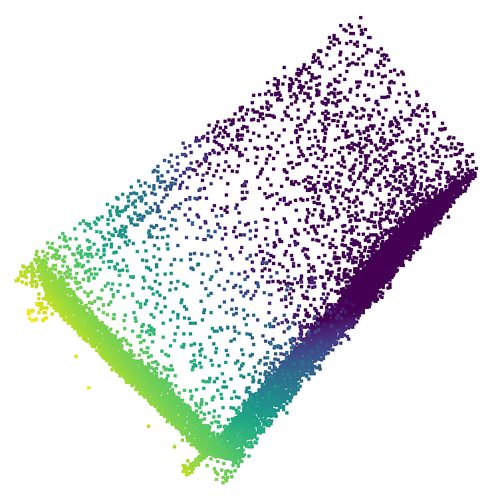} & 
        \includegraphics[width=30pt]{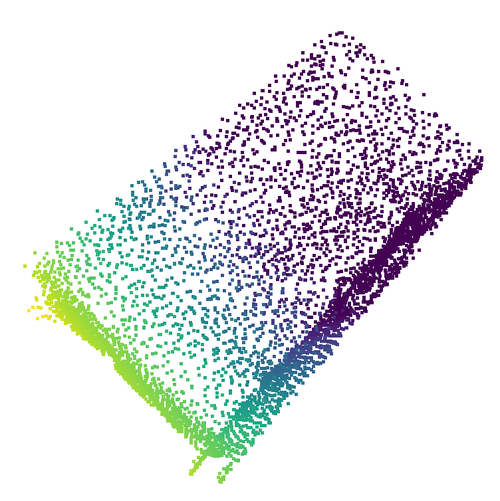} & 
        \includegraphics[width=30pt]{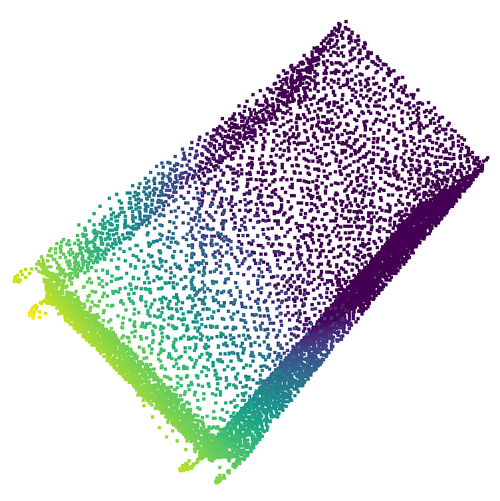} & 
        \includegraphics[width=30pt]{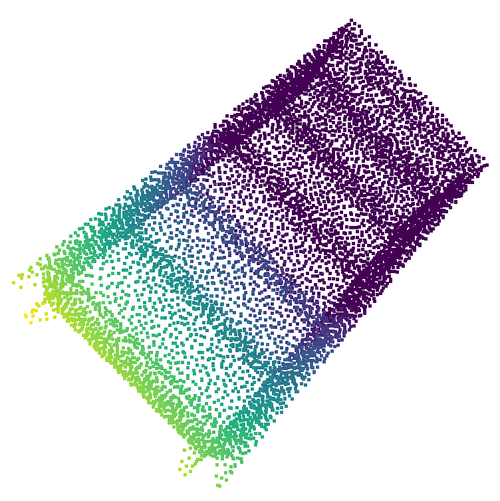} \\

        \includegraphics[width=30pt]{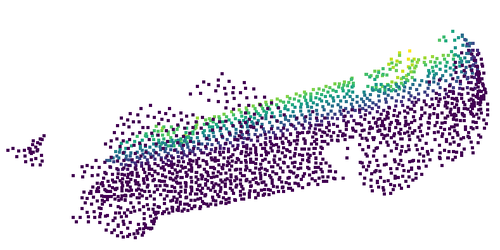} & 
        \includegraphics[width=30pt]{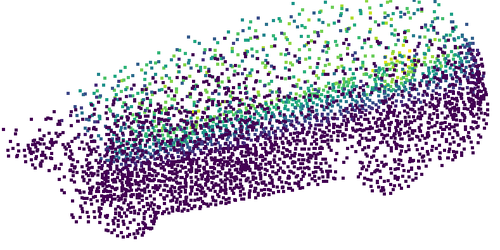} & 
        \includegraphics[width=30pt]{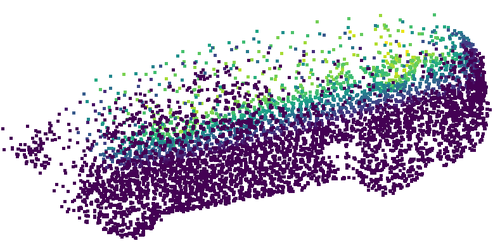} & 
        \includegraphics[width=30pt]{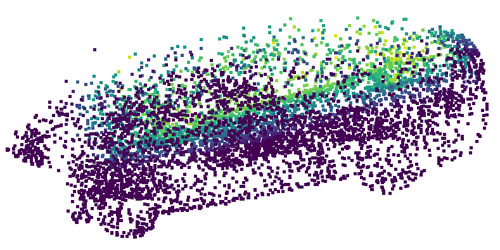} & 
        \includegraphics[width=30pt]{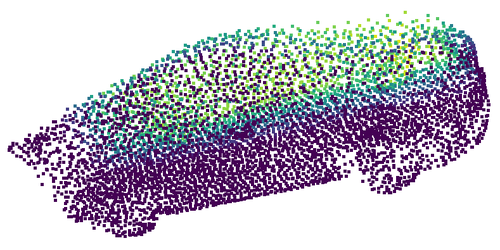} & 
        \includegraphics[width=30pt]{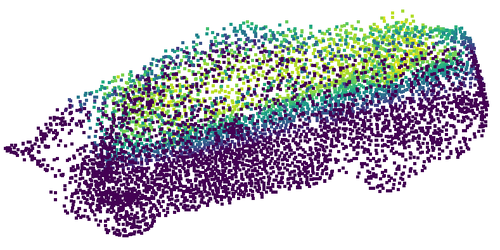} & 
        \includegraphics[width=30pt]{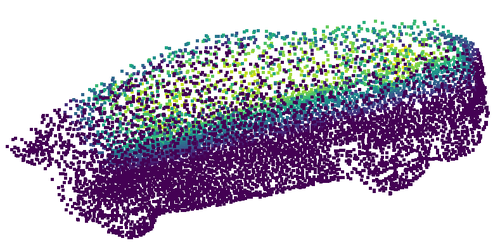} & 
        \includegraphics[width=30pt]{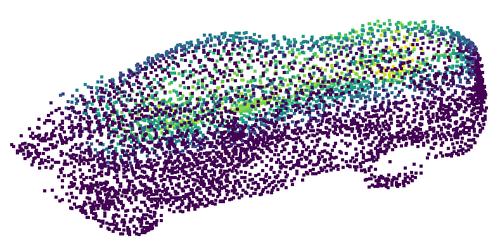} & 
        \includegraphics[width=30pt]{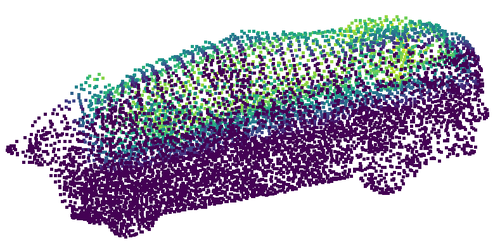} & 
        \includegraphics[width=30pt]{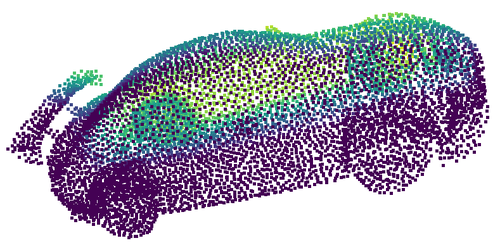} \\

        \includegraphics[width=30pt]{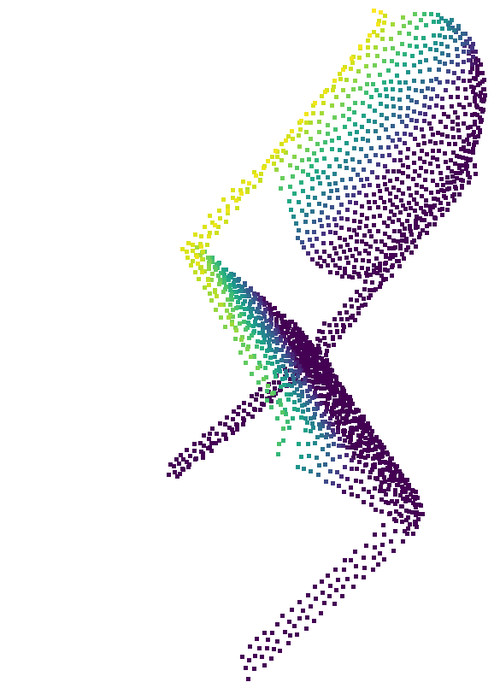} & 
        \includegraphics[width=30pt]{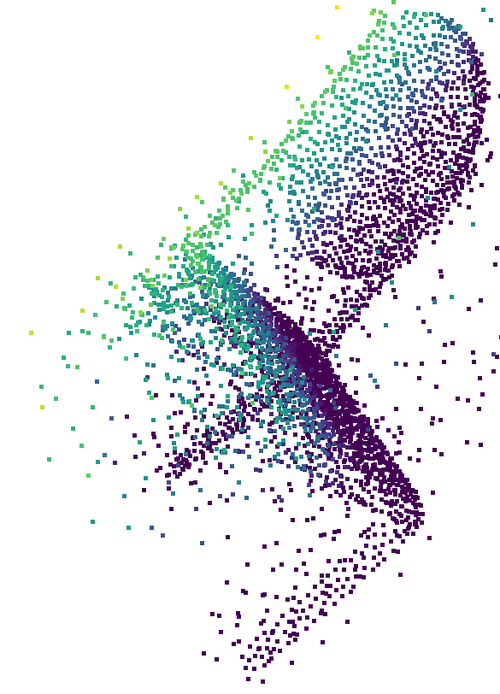} & 
        \includegraphics[width=30pt]{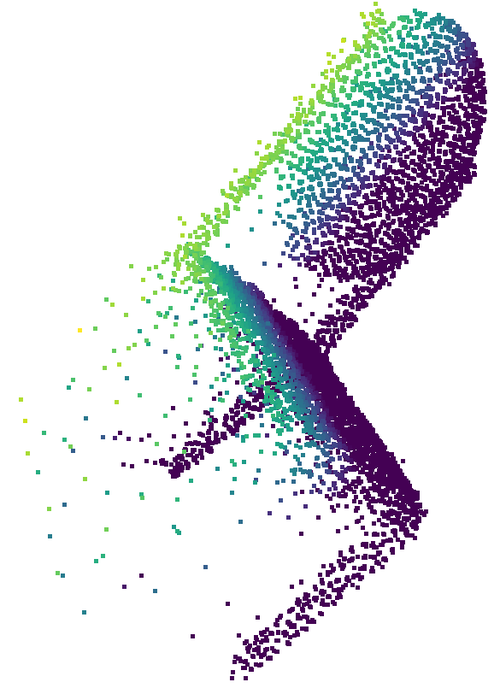} & 
        \includegraphics[width=30pt]{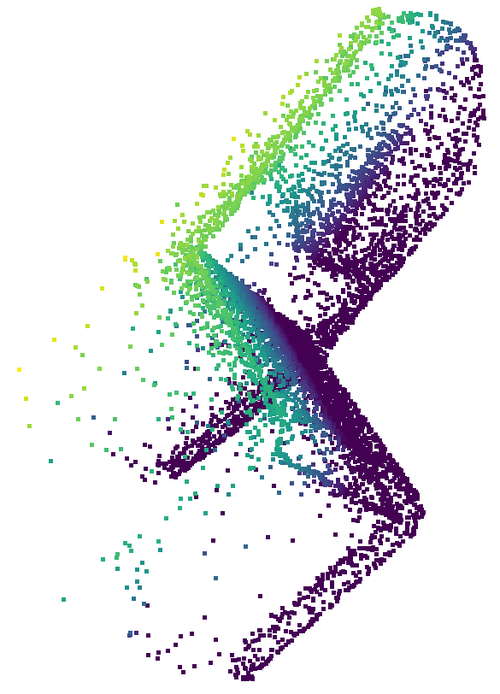} & 
        \includegraphics[width=30pt]{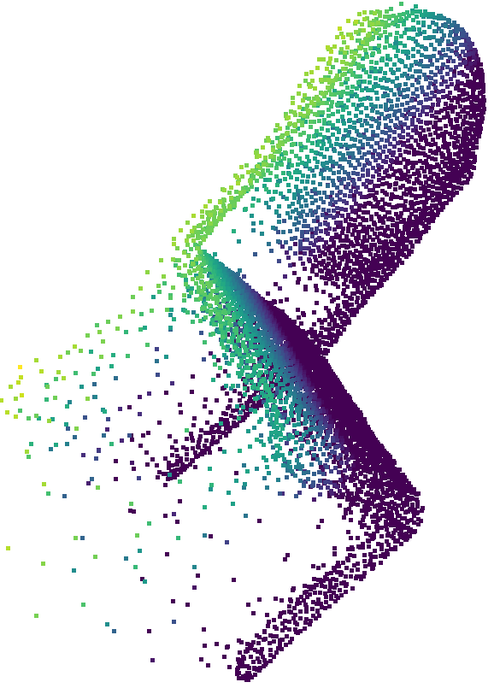} & 
        \includegraphics[width=30pt]{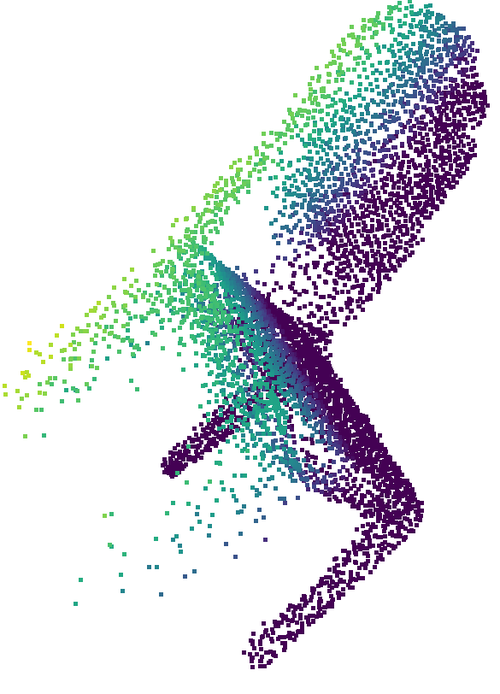} & 
        \includegraphics[width=30pt]{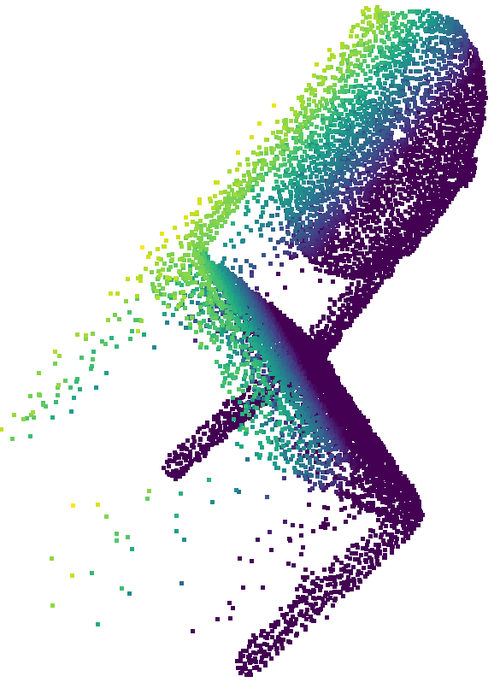} & 
        \includegraphics[width=30pt]{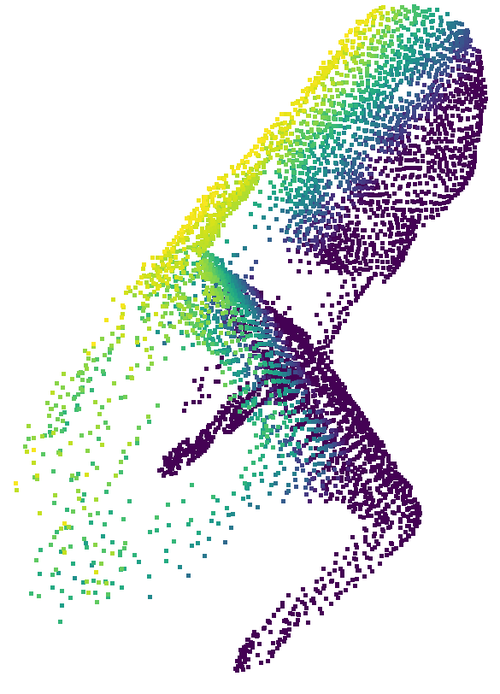} & 
        \includegraphics[width=30pt]{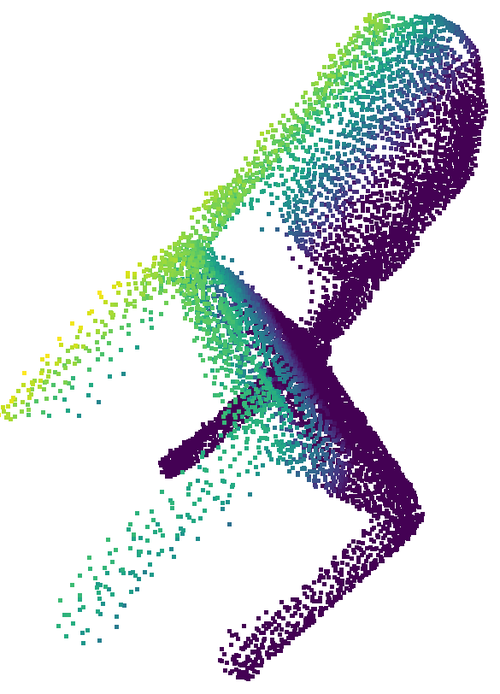} & 
        \includegraphics[width=30pt]{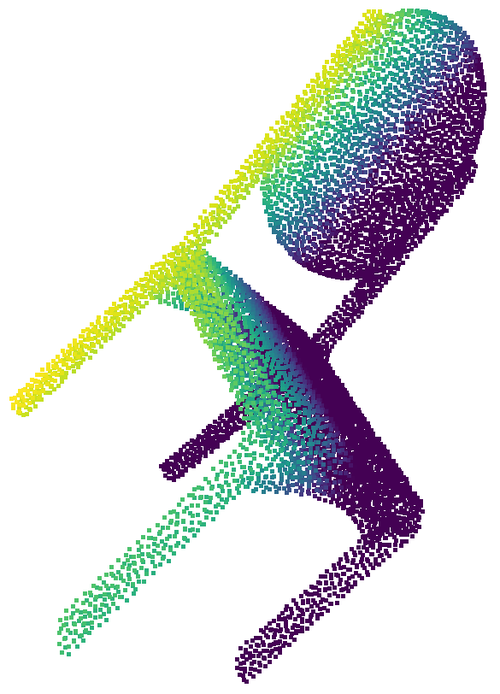} \\

        \includegraphics[width=30pt]{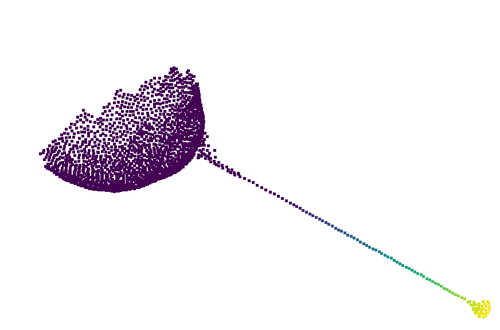} & 
        \includegraphics[width=30pt]{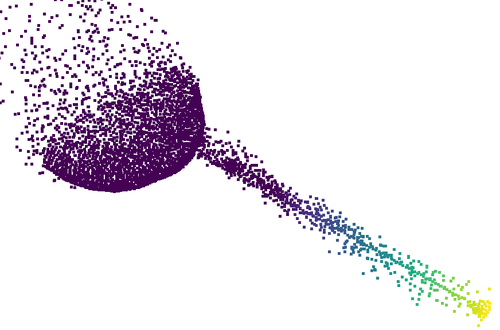} & 
        \includegraphics[width=30pt]{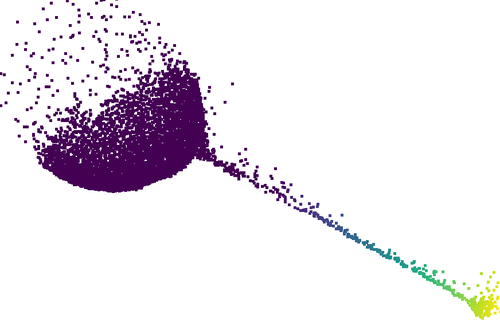} & 
        \includegraphics[width=30pt]{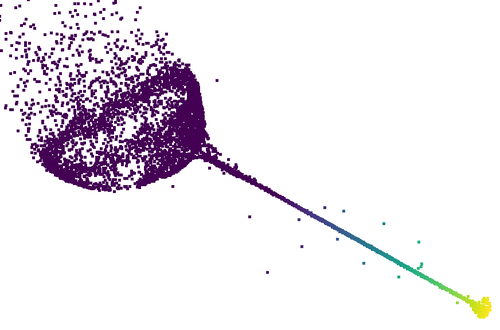} & 
        \includegraphics[width=30pt]{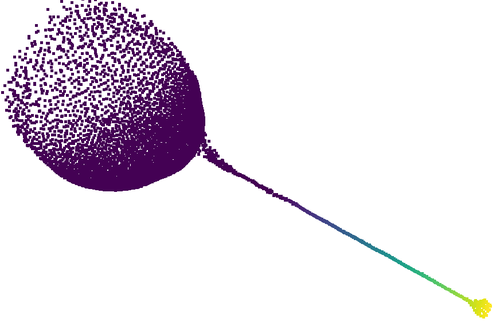} & 
        \includegraphics[width=30pt]{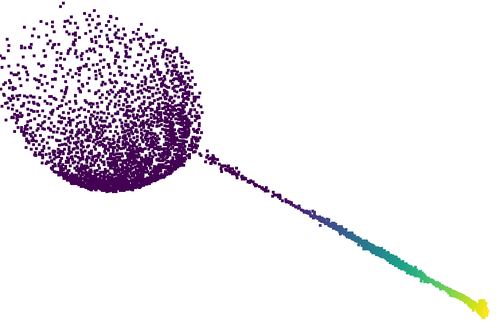} & 
        \includegraphics[width=30pt]{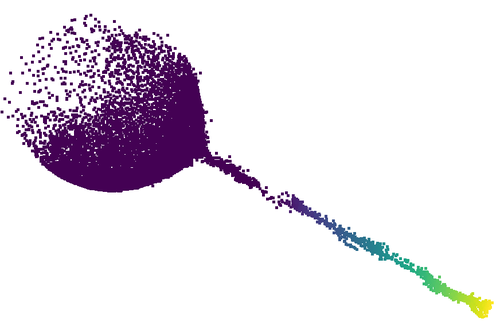} & 
        \includegraphics[width=30pt]{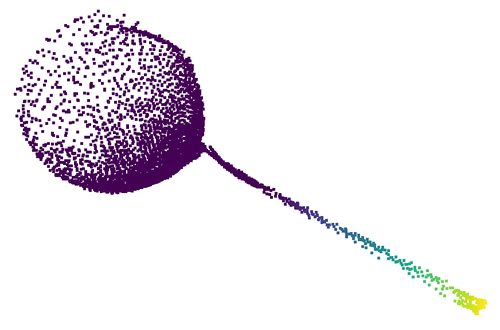} & 
        \includegraphics[width=30pt]{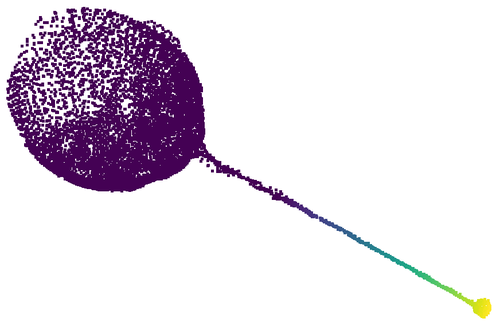} & 
        \includegraphics[width=30pt]{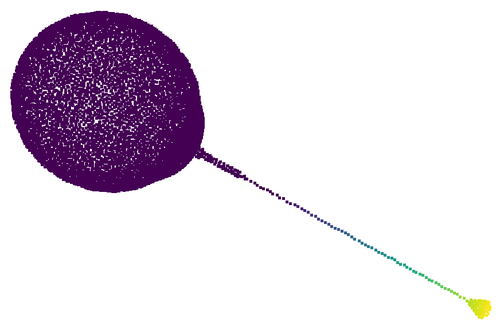} \\

        \includegraphics[width=30pt]{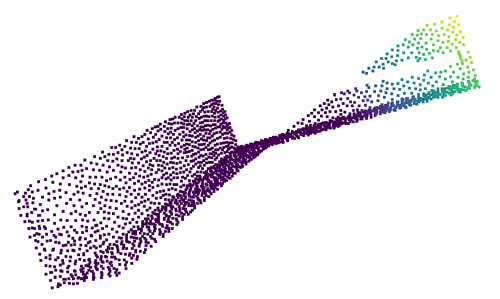} & 
        \includegraphics[width=30pt]{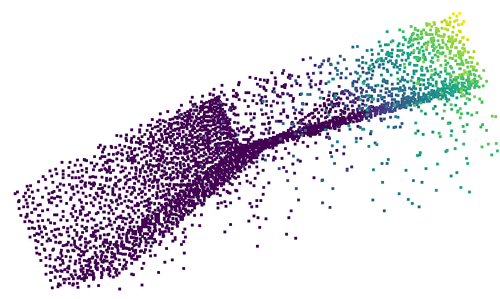} & 
        \includegraphics[width=30pt]{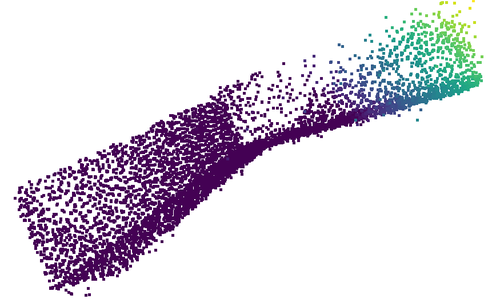} & 
        \includegraphics[width=30pt]{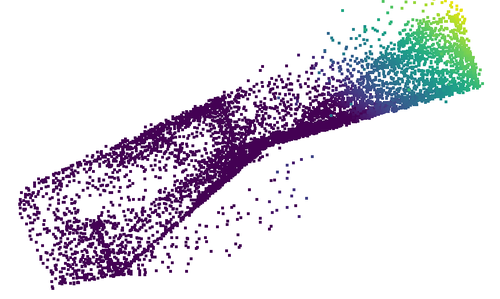} & 
        \includegraphics[width=30pt]{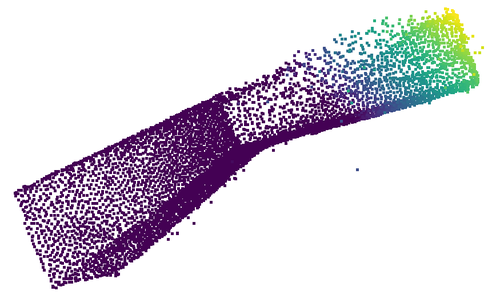} & 
        \includegraphics[width=30pt]{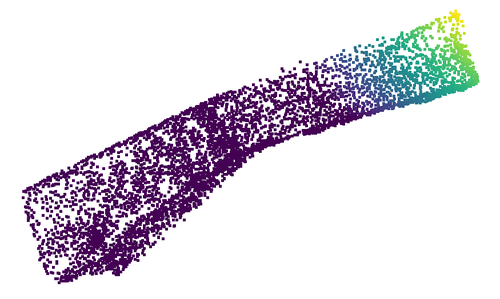} & 
        \includegraphics[width=30pt]{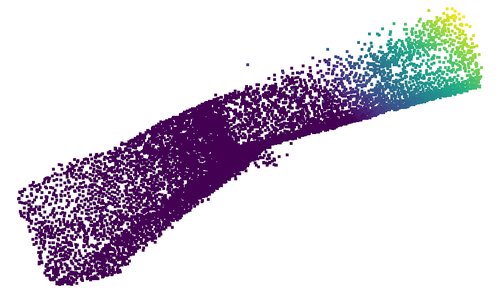} & 
        \includegraphics[width=30pt]{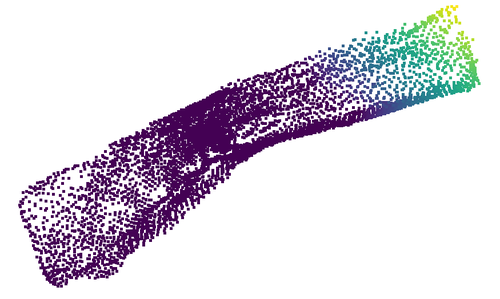} & 
        \includegraphics[width=30pt]{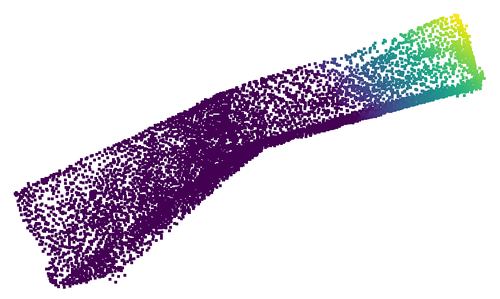} & 
        \includegraphics[width=30pt]{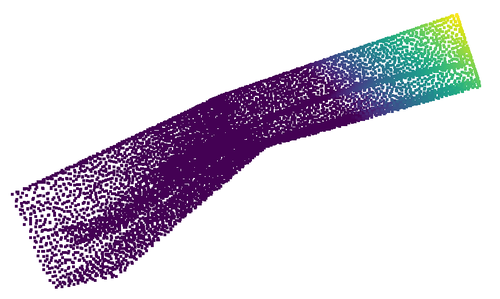} \\

        \includegraphics[width=30pt]{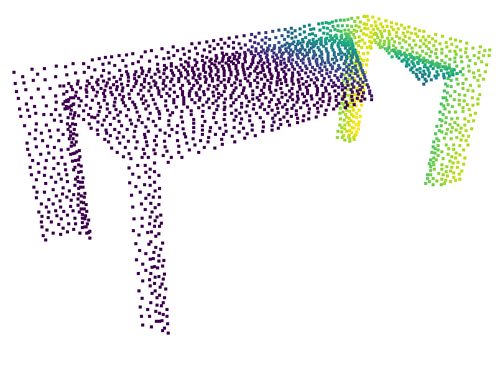} & 
        \includegraphics[width=30pt]{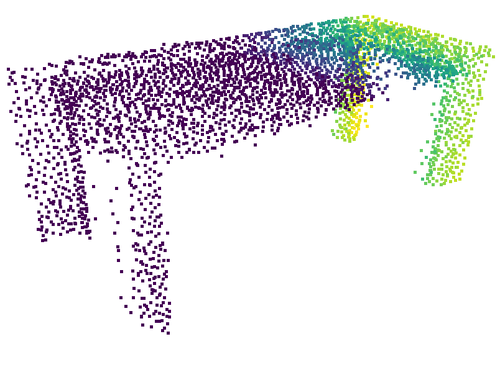} & 
        \includegraphics[width=30pt]{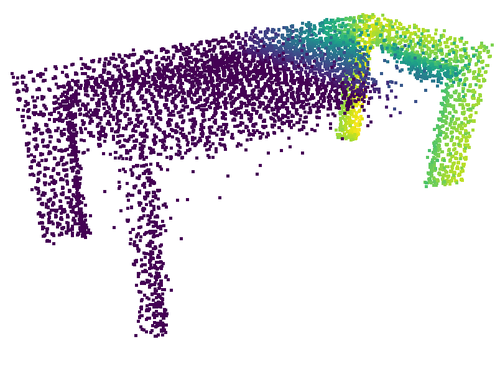} & 
        \includegraphics[width=30pt]{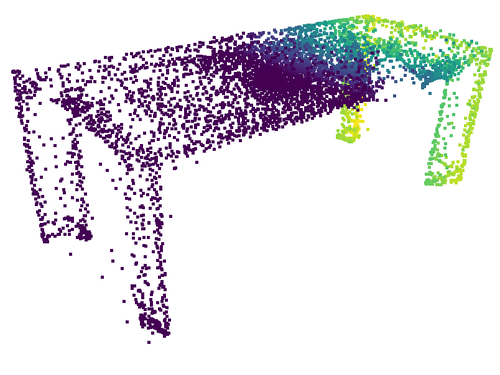} & 
        \includegraphics[width=30pt]{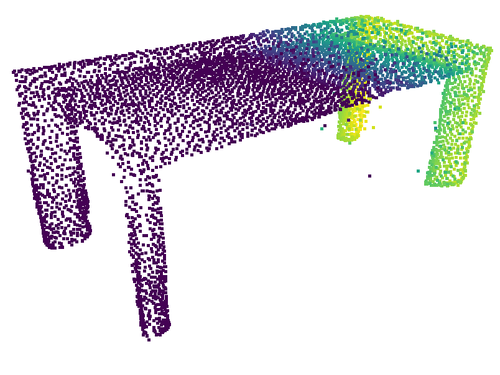} & 
        \includegraphics[width=30pt]{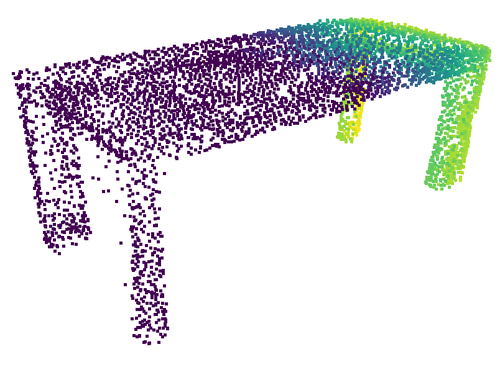} & 
        \includegraphics[width=30pt]{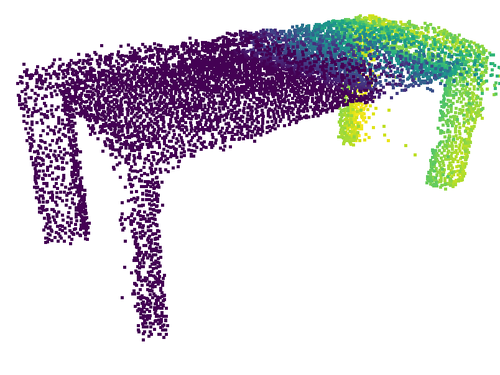} & 
        \includegraphics[width=30pt]{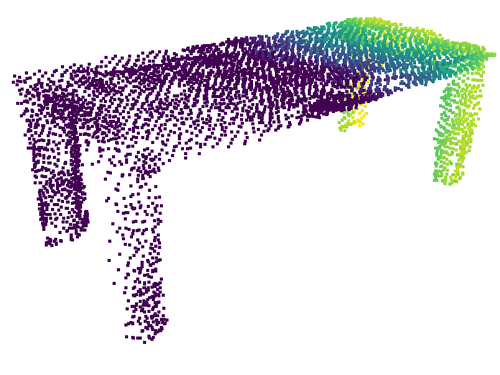} & 
        \includegraphics[width=30pt]{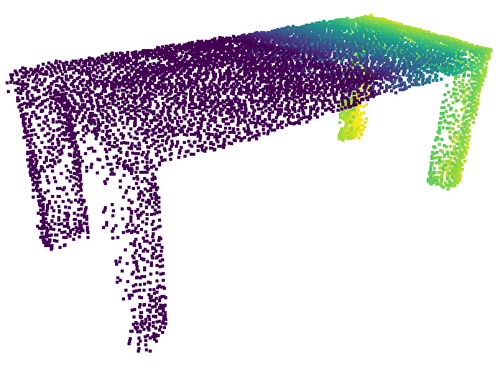} & 
        \includegraphics[width=30pt]{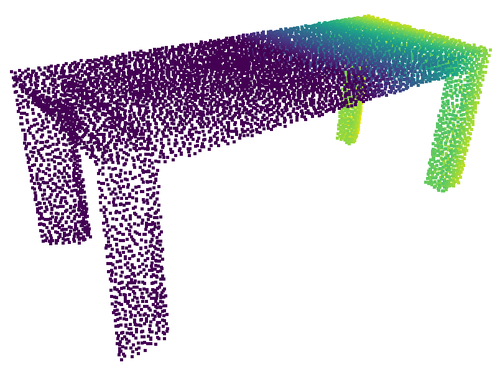} \\

        \includegraphics[width=30pt]{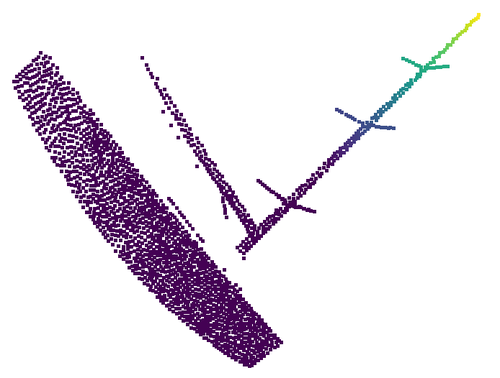} & 
        \includegraphics[width=30pt]{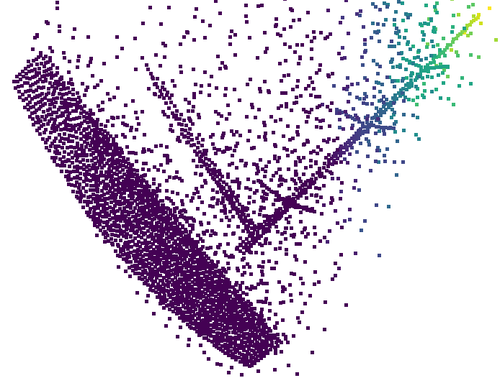} & 
        \includegraphics[width=30pt]{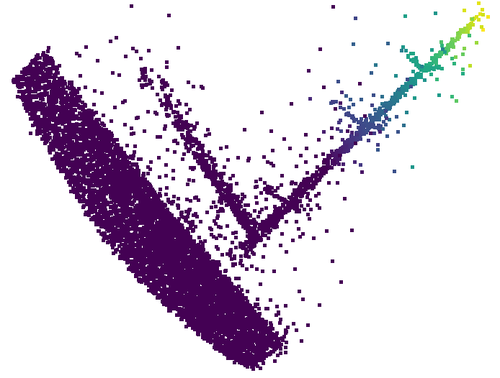} & 
        \includegraphics[width=30pt]{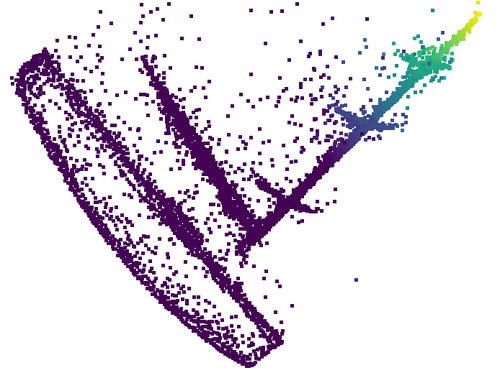} & 
        \includegraphics[width=30pt]{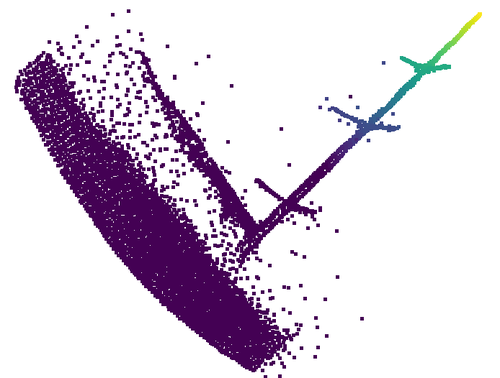} & 
        \includegraphics[width=30pt]{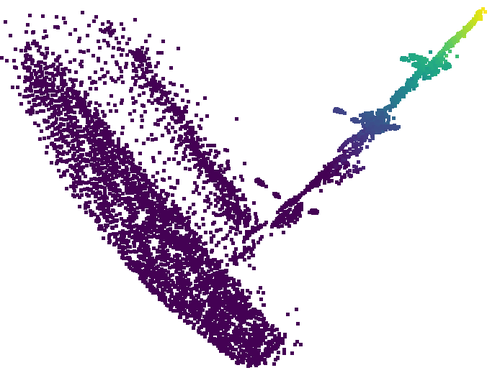} & 
        \includegraphics[width=30pt]{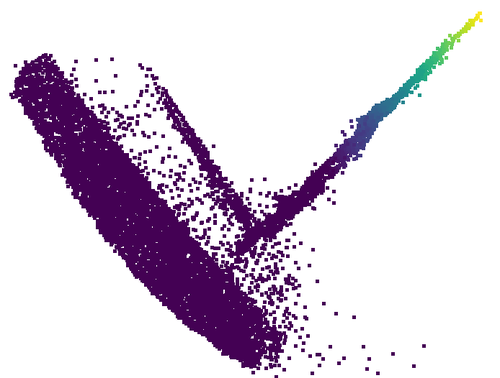} & 
        \includegraphics[width=30pt]{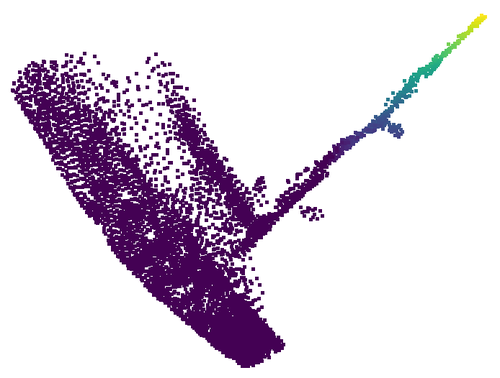} & 
        \includegraphics[width=30pt]{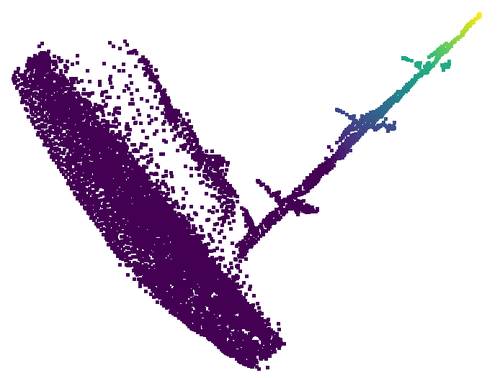} & 
        \includegraphics[width=30pt]{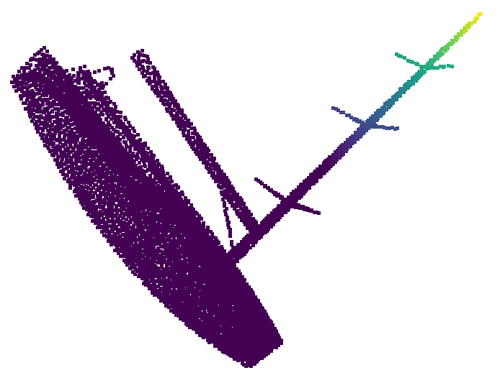} \\
    \end{tabular}
    \caption{SO(3)-equivariant point cloud completion results on MVP dataset.}
    \label{fig:qualitative}
\end{figure}

By inspecting the generated point clouds on the MVP dataset (Fig.~\ref{fig:qualitative}), we observe that our anchor-based framework can better preserve and reconstruct local geometric details compared to EquivPCN~\cite{comp_equiv:equiv_pcn}. In addition, most rotation-variant methods emphasize accurately reconstructing the observed region to achieve lower CD-$l_1$, but often generate noisy or incoherent structures for the unobserved parts. ODGNet and AdaPoinTr shows improvements in producing more detailed completions, yet outliers and structural artifacts are still noticeable.

\setlength{\tabcolsep}{1.5pt}

We then evaluate our model's generalizability to real-world point clouds using the KITTI dataset under None/None setup. As shown in Table~\ref{tab:kitti_metrics}, our framework achieves a relatively low FD score, indicating high fidelity to the observed input geometry. For MMD, REVNET performs comparably to global-feature-based, non-equivariant methods. We further report the performance of our method when the input point cloud contains more than 30 points, where the MMD score improves significantly. Although a performance gap remains compared to recent rotation-variant models, it is important to note that these methods rely on aligned inputs using hand-labeled bounding boxes from the KITTI benchmark. In contrast, our method can directly operate on arbitrarily oriented inputs, avoiding the potential errors and noise introduced by bounding box detection in real-world deployment.

\setlength{\tabcolsep}{2pt}
\begin{table}
    \centering
    \small
    \begin{tabular}{c|c c c c c c c c|c}
        \hline
        Method & Air & Cab & Car & Cha & Lam & Sof & Tab & Ves & Avg \\
        \hline
        & \multicolumn{9}{c}{SO(3)/SO(3)} \\
        \hline
        PCN & 0.29 & 0.40 & 0.46 & 0.49 & 0.60 & 0.48 & 0.62 & 0.52 & 0.48 \\
        TopNet & 0.36 & 0.49 & 0.46 & 0.52 & 0.78 & 0.50 & 0.67 & 0.69 & 0.55 \\
        MSN & 0.32 & 0.39 & 0.35 & 0.48 & 0.51 & 0.40 & 0.58 & 0.47 & 0.44 \\
        CRN & 0.21 & 0.22 & 0.16 & 0.36 & 0.31 & 0.27 & 0.37 & 0.24 & 0.27 \\
        GRNet & 0.18 & 0.28 & 0.19 & 0.30 & 0.26 & 0.29 & 0.43 & 0.31 & 0.28 \\
        VRCNet & 0.13 & 0.16 & 0.11 & 0.23 & 0.21 & 0.21 & 0.29 & 0.28 & 0.20 \\
        SnowFlake & 0.12 & 0.20 & 0.11 & 0.27 & 0.24 & 0.25 & 0.31 & 0.28 & 0.22 \\
        PoinTr & 0.54 & 0.32 & 0.28 & 0.47 & 0.44 & 0.40 & 0.54 & 0.49 & 0.44 \\
        AnchorFormer & 0.34 & 1.29 & 2.12 & 1.41 & 2.30 & 2.05 & 1.67 & 1.98 & 1.65 \\
        PMP-Net++ & 0.08 & 0.14 & 0.09 & 0.17 & 0.15 & 0.14 & 0.18 & 0.16 & 0.14 \\
        GTNet & 0.10 & 0.17 & 0.11 & 0.20 & 0.18 & 0.17 & 0.21 & 0.19 & 0.16 \\
        ODGNet & 0.20 & 0.24 & 0.18 & 0.31 & 0.35 & 0.24 & 0.37 & 0.29 & 0.27 \\
        AdaPoinTr & 0.13 & 0.22 & 0.14 & 0.22 & 0.22 & 0.22 & 0.23 & 0.20 & 0.20 \\
        \hline
        & \multicolumn{9}{c}{None/SO(3)} \\
        \hline
        EquivPCN & 0.06 & 0.12 & 0.08 & 0.13 & 0.14 & \textbf{0.11} & 0.15 & 0.11 & 0.11 \\
        ESCAPE & 0.10 & 0.18 & 0.13 & 0.18 & 0.14 & 0.17 & 0.17 & 0.16 & 0.15 \\
        Ours    & \textbf{0.04} & \textbf{0.10} & \textbf{0.05} & \textbf{0.11} & \textbf{0.10} & \textbf{0.11} & \textbf{0.12} & \textbf{0.10} & \textbf{0.09} \\
        \hline
    \end{tabular}
    \caption{Completion consistency measured by $cst_{\text{CD-}l_1} \times 100 \downarrow$ on the MVP dataset with a resolution of 8192 points. The best results are highlighted in bold. Our model produces highly consistent completions under arbitrary rotations. }
    \label{tab:mvp_cst}
\end{table}

\begin{figure}
    \centering
    \small
    \renewcommand{\arraystretch}{0.0}
    \begin{tabular}{c m{44pt}<{\centering} m{44pt}<{\centering} m{44pt}<{\centering} m{44pt}<{\centering} m{44pt}<{\centering} m{44pt}<{\centering}}
        Input & 
        \includegraphics[width=44pt]{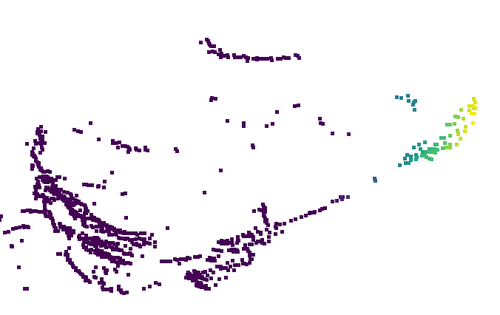} & 
        \includegraphics[width=44pt]{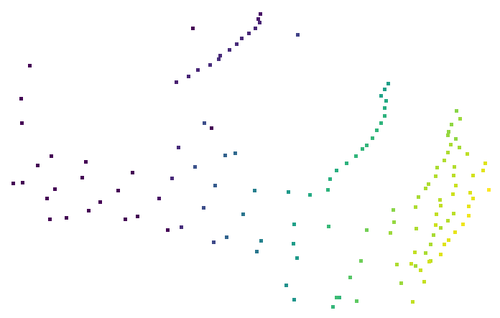} & 
        \includegraphics[width=44pt]{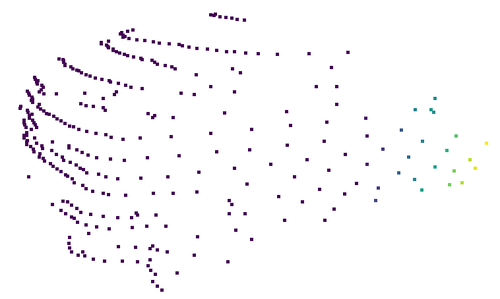} & 
        \includegraphics[width=44pt]{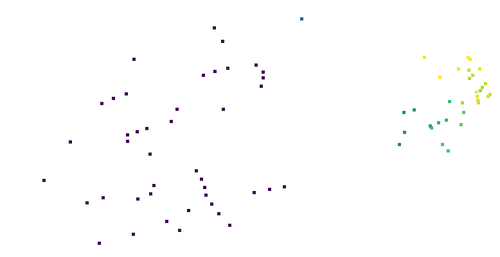} & 
        \includegraphics[width=44pt]{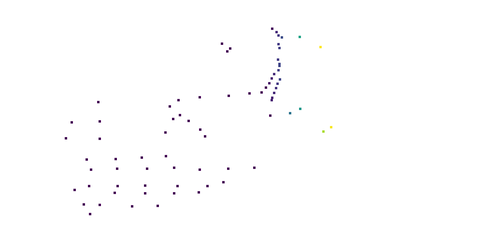} & 
        \includegraphics[width=44pt]{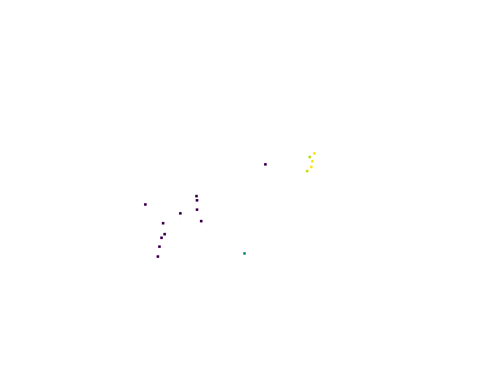} \\

        PoinTr & 
        \includegraphics[width=44pt]{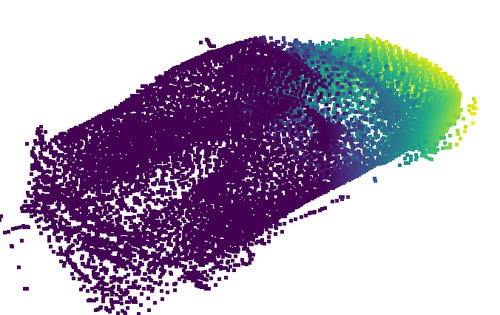} & 
        \includegraphics[width=44pt]{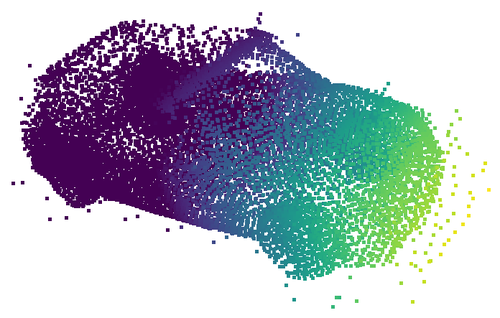} & 
        \includegraphics[width=44pt]{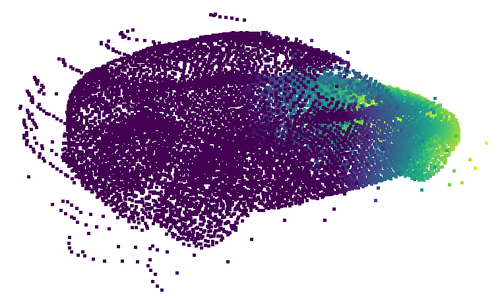} & 
        \includegraphics[width=44pt]{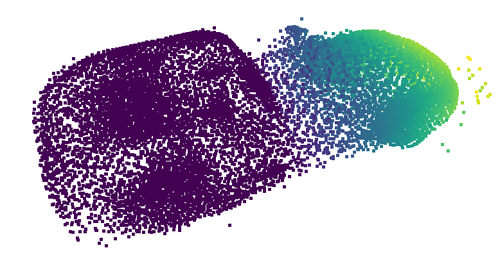} & 
        \includegraphics[width=44pt]{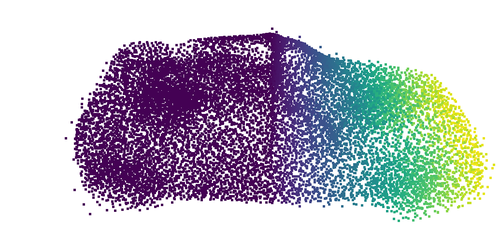} & 
        \includegraphics[width=44pt]{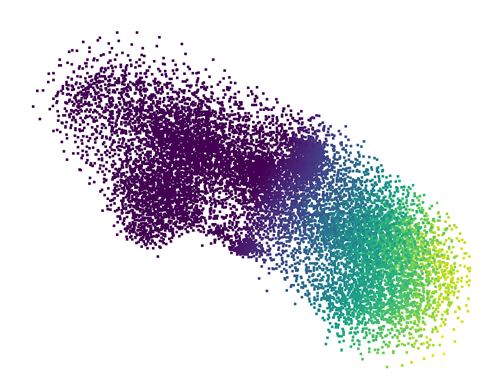} \\

        ESCAPE & 
        \includegraphics[width=44pt]{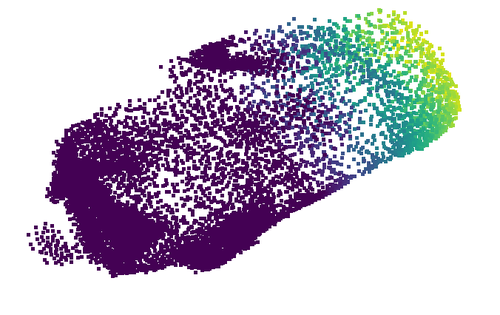} & 
        \includegraphics[width=44pt]{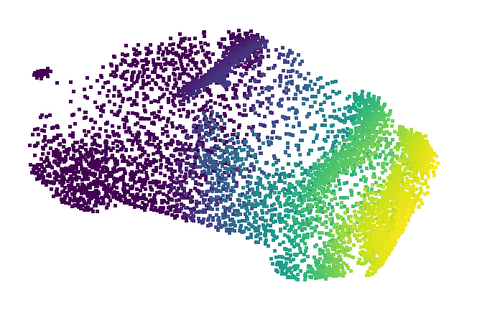} & 
        \includegraphics[width=44pt]{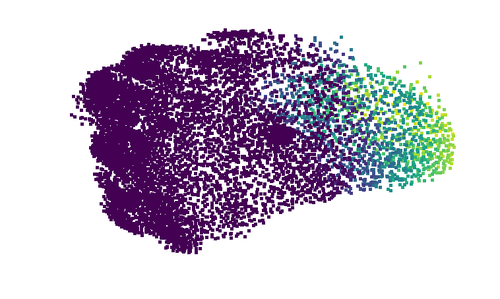} & 
        \includegraphics[width=44pt]{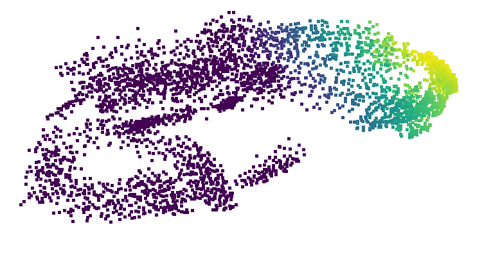} & 
        \includegraphics[width=44pt]{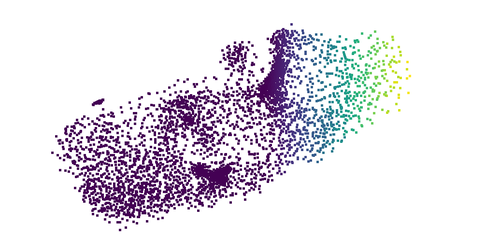} & 
        \includegraphics[width=44pt]{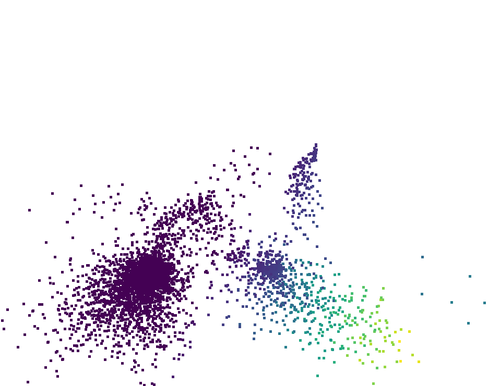} \\

        Ours & 
        \includegraphics[width=44pt]{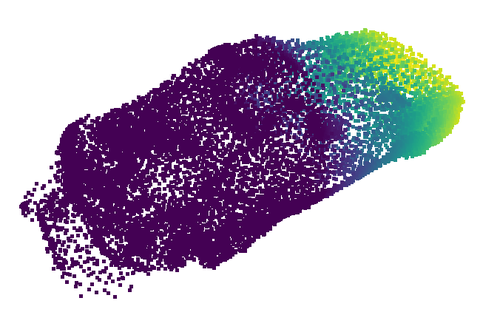} & 
        \includegraphics[width=44pt]{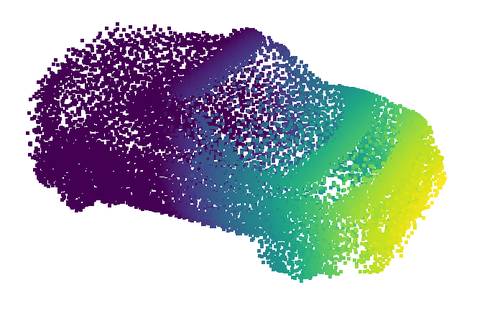} & 
        \includegraphics[width=44pt]{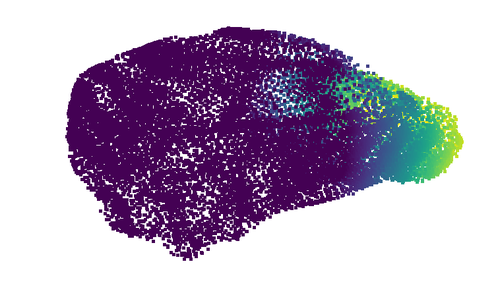} & 
        \includegraphics[width=44pt]{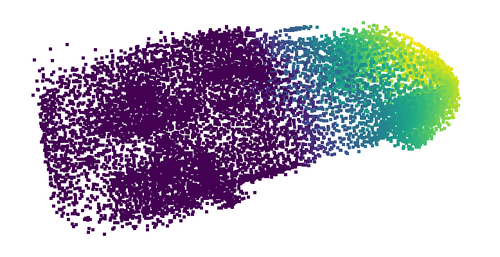} & 
        \includegraphics[width=44pt]{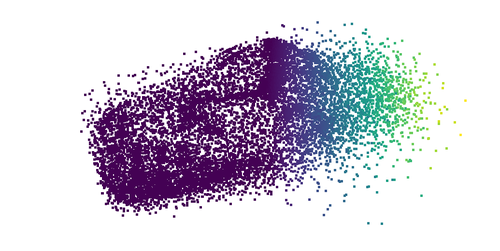} & 
        \includegraphics[width=44pt]{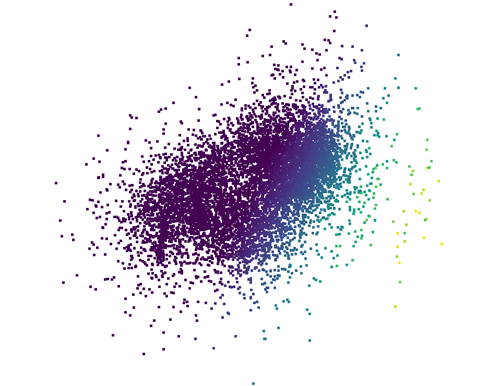} \\

    \end{tabular}
    \caption{Car point cloud completion results on KITTI dataset.}
    \label{fig:kitti}
\end{figure}

Qualitative results on KITTI are shown in Fig.~\ref{fig:kitti}. With sufficient input points, REVNET produces clean and reliable car completions that are well aligned with the observed geometry. However, when the input is extremely sparse ($<10$ points), both rotation-equivariant models fail to infer a reasonable structure, whereas PoinTr can still generate a coarse car shape. We believe this stems from a fundamental difference between rotation-variant and equivariant models: rotation-variant methods trained in canonical pose only need to generate a canonical car shape, while an equivariant model must also infer the implicit pose information from the observation, which becomes unstable when too few cues are available. In addition, the MMD metric can be affected by imperfect bounding-box alignment during evaluation, which introduces pose discrepancies between equivariant predictions and canonical reference models. This suggests that future benchmark for equivariant point cloud completion on real-world data should incorporate mechanisms to handle these perturbations more robustly.

\setlength{\tabcolsep}{0.2pt}
\begin{table}
    \centering
    \begin{tabular}{c|m{30pt}<{\centering} m{30pt}<{\centering} m{30pt}<{\centering} m{30pt}<{\centering} m{30pt}<{\centering} m{30pt}<{\centering} m{30pt}<{\centering} m{30pt}<{\centering} | m{30pt}<{\centering} m{40pt}<{\centering}}
        \hline
         & PCN & TopNet & MSN & PoinTr & GTNet & ODG-Net & Ada-PoinTr & ESC-APE & Ours & Ours $>30$ pts\\
        \hline
        FD$\downarrow$     & 2.235 & 5.354 & 0.434 & \textbf{0.000} & 0.018 & 1.280 & 0.237 & 1.416 & 0.612 & 0.741 \\
        MMD$\downarrow$    & 1.366 & 0.636 & 2.259 & 0.526 & 0.353 & \textbf{0.349} & 0.392 & 4.420 & 0.771 & 0.657 \\
        \hline
    \end{tabular}
    \caption{Performance comparison in terms of FD $\times 1000 \downarrow$ and MMD $\times 1000 \downarrow$ on the KITTI dataset under None/None setup. Our model achieves competitive results to global-feature-based rotation-variant models while not requiring a rotation alignment using hand-labeled bounding boxes.}
    \label{tab:kitti_metrics}
\end{table}

\setlength{\tabcolsep}{3pt}

\setlength{\tabcolsep}{5pt}
\begin{table}
    \centering
    \small
    \begin{tabular}{c|c|c c c c c c}
        \hline
         & Ours & (1) & (2) & (3) & (4) & (5) & (6) \\
        \hline
        CD-$l_1 \times 100 \downarrow$ & \textbf{0.887} & 0.914 & 0.927 & 0.903 & 0.911 & 0.918 & 0.905 \\
        F-Score@1$\%\uparrow$ & \textbf{73.44} & 71.60 & 70.20 & 72.24 & 71.49 & 71.32 & 72.74 \\
        F-Score@2$\%\uparrow$ & \textbf{92.37} & 91.77 & 91.28 & 91.87 & 91.74 & 91.54 & 91.97\\
        \hline
    \end{tabular}
    \caption{Ablation study on our framework on the MVP dataset using the resolution of 8192 points. In each experiment, we only replace one component with another approach.}
    \label{tab:ablation}
\end{table}

\subsection{Ablation Study}

We conduct an extensive ablation study to evaluate the contribution of each proposed component. Each experiment replaces a single module with an alternative design: (1) the proposed feature backbone is replaced with a VN-DGCNN. (2) The missing anchor predictor and local invariant dense decoder are replaced with the DSK module and VN-Folding decoder from~\cite{comp_equiv:equiv_pcn}. (3) The channel-wise subtraction attention in the Transformer is replaced with Frobenius inner product attention~\cite{rotequiv:vntransformer}. (4) The normalization layers in the backbone are replaced with VN-BatchNorm~\cite{rotequiv:vnn}. (5) All VN-ZCALayerNorm are replaced with the 2-norm VN-LayerNorm~\cite{rotequiv:vntransformer}. (6) All VN-bias terms are omitted. As shown in Table~\ref{tab:ablation}, all proposed components consistently improve the performance. We hope these findings can inspire more advanced design of VN networks in future work.

\section{Conclusion}
\label{sec:conclusion}

In this paper, we present a VN-based framework to recover complete point clouds from partial observations in arbitrary poses. We introduce a rotation-equivariant backbone to extract representative anchor features and propose the VN Missing Anchor Transformer, which employs channel-wise subtraction attention to infer equivariant features for missing anchors. By leveraging the flexible equivariant-to-invariant conversion property, our model generates point coordinates from VN features with greater stability. The proposed framework outperforms all baselines on equivariant completion on synthetic dataset and delivers competitive results on real-world dataset with pose-aligned input. Future work includes integrating our approach into a scene completion scheme and exploring its potential in unsupervised point cloud completion.

\subsubsection{Acknowledgements} This work has been supported by the German Federal Ministry of Research, Technology and Space (BMFTR) under the Robotics Institute Germany (RIG) and the Bavarian Ministry of Economic Affairs, Regional Development and Energy as part of the project ``6G Future Lab Bavaria''.

%
%
%
\bibliographystyle{splncs04}
\bibliography{main}

\title{Supplementary Material\\REVNET: Rotation-Equivariant Point Cloud Completion via Vector Neuron
Anchor Transformer}
\titlerunning{Supplementary Material for REVNET}
%
\author{Zhifan Ni\inst{1} \and
Eckehard Steinbach\inst{1}}
\authorrunning{Z. Ni and E. Steinbach}
%
\institute{Technical University of Munich, Munich, Germany \\
\email{\{zhifan.ni, eckehard.steinbach\}@tum.de}}
\maketitle              
\section{Vector Neuron Network}
\subsection{Core Layers}
Vector Neuron (VN) Network~\cite{rotequiv:vnn} represents a latent feature as an ordered list of 3D vectors $\mathbf{X} \in \mathbb{R}^{C \times 3}$, where $C$ denotes the number of channels. This representation enables rotation of the input to propagate to deeper network layers, which means that applying a SO(3) transformation $\mathbf{R}^{3 \times 3}$ to the input will also transform the output in the same way. Following layers are introduced in~\cite{rotequiv:vnn} to build essential components of a neural network.

\subsubsection{Linear Layer.}
The linear layer is defined as $\text{VN-Linear}(\mathbf{X};\mathbf{W}) = \mathbf{W}\mathbf{X} \in \mathbb{R}^{C^\prime \times 3}$, where $\mathbf{W} \in \mathbb{R}^{C^\prime \times C}$ is a learnable weight matrix. Applying rotation to the input leads to:
$$\text{VN-Linear}(\mathbf{X}\mathbf{R};\mathbf{W}) = \mathbf{W}\mathbf{X}\mathbf{R} = \text{VN-Linear}(\mathbf{X};\mathbf{W})\mathbf{R}\text{,}$$
which shows the equivariance. 

\subsubsection{Non-linear Layer.}
Non-linear layers are crucial to the representation power of a neural network. A VN-ReLU layer~\cite{rotequiv:vnn} learns a weight matrix to map a neuron $\mathbf{x} \in \mathbb{R}^{1 \times 3}$ to a direction $\mathbf{k} \in \mathbb{R}^{1 \times 3}$. If the original neuron follows the direction of $\mathbf{k}$, i.e., $\langle\mathbf{x},\mathbf{k}\rangle > 0$, $\mathbf{x}$ remains unchanged. Otherwise, the output is the projection of $\mathbf{x}$ onto the plane orthogonal to $\mathbf{k}$: $\mathbf{x} - \langle\mathbf{x},\frac{\mathbf{k}}{\Vert\mathbf{k}\Vert}\rangle\frac{\mathbf{k}}{\Vert\mathbf{k}\Vert}$. Because the inner-product is invariant to rotation $\langle\mathbf{x}\mathbf{R},\mathbf{k}\mathbf{R}\rangle = \mathbf{x}\mathbf{R}\mathbf{R}^\top\mathbf{k}^\top = \mathbf{x}\mathbf{k}^\top = \langle\mathbf{x}, \mathbf{k}\rangle$, the VN-ReLU layer is rotation-equivariant.

\subsubsection{Pooling Layer.}
Pooling layers are essential for information aggregation. In VN networks, similar to the VN-ReLU, the VN-MaxPooling layer~\cite{rotequiv:vnn} learns a weight to map neurons to a direction and select the neuron that best aligns with it. However, the selection operation blocks the gradient propagation to the learnable direction mapping, making the VN-MaxPooling layer non-trainable. As also observed in the original VN paper~\cite{rotequiv:vnn}, average pooling consistently outperforms VN-MaxPooling. Therefore, we adopt average pooling throughout our framework.

\subsubsection{Batch Normalization.}
Normalization layers are commonly used to stabilize the network training and improve performance. VN-BatchNorm~\cite{rotequiv:vnn} extends this concept to VN networks by normalizing the rotation-invariant 2-norm of VN features across a batch. However, we observe that VN-BatchNorm can occasionally degrade performance compared to no normalization at all. This motivates us to explore more effective normalization strategies for VN frameworks. 

\subsubsection{Multi-layer Perception (MLP).}
A VN-MLP block consists of a series of VN-Linear, VN-ReLU, and VN-Normalization layers. One can add the output of a VN-MLP to its input to build a residual VN-MLP block. 

\subsubsection{Edge Convolution (EdgeConv).}
EdgeConv, introduced in DGCNN~\cite{3d_point:dgcnn}, aggregates local geometric information by constructing a dynamic graph via k-nearest neighbors. It computes edge features based on point relationships, updates them using an MLP, and fuses the results through pooling operations. Replacing the conventional components with corresponding VN layers yields the rotation-equivariant VN-EdgeConv block. In our framework, we adopt this block as a point coordinate embedding module to lift the 1-channel 3D coordinate $\mathbf{p} \in 
\mathbb{R}^{1 \times 3}$ into a high-dimensional VN space. 

\subsubsection{Invariant Layer.}
The VN features can be easily transformed to rotation-invariant features. The VN-Inv layer~\cite{rotequiv:vnn} uses a VN-MLP block to produce a transformation matrix $\mathbf{T} = \text{VN-MLP}(\mathbf{X}) \in \mathbb{R}^{3 \times 3}$ and derive the invariant feature as $\text{VN-Inv}(\mathbf{X}) = \mathbf{X} \mathbf{T}^\top$. This $C \times 3$ feature can be flattened to a vector and fed to conventional neural network components. Afterwards, the new scalar feature $\mathbf{x}^\prime_{\text{inv}} \in \mathbb{R}^{3C^\prime}$ can be reshaped to $\mathbf{X}^\prime_{\text{inv}} \in \mathbb{R}^{C^\prime \times 3}$ and transformed back to the input frame using the inverse transformation matrix: $\mathbf{X}^\prime = \mathbf{X}^\prime_{\text{inv}}(\mathbf{T}^\top)^{-1}$. By forcing the transformation matrix to be orthonormal, we can simply apply $\mathbf{X}^\prime = \mathbf{X}^\prime_{\text{inv}}\mathbf{T}$. We typically adopt the VN-Inv layers with orthonormal transformation matrix to convert VN features for generating 3D point coordinates. The rotation-invariant point clouds are then transformed back to the frame of the input. We also use VN-Inv layers to obtain rotation-invariant attention score in the channel-wise subtraction attention.

\subsection{VN-Bias}

On the KITTI dataset, we observe that VN-based point cloud completion methods without a bias term perform extremely poorly when the number of input points is very low (typically fewer than 30), whereas rotation-variant models such as PoinTr~\cite{comp:pointr} can still generate a plausible car shape. We hypothesize that one major contributing factor to this failure in sparse settings is the absence of bias terms in VN layers. When the input contains very few points, the farthest point sampling step often yields duplicated anchor positions. Consequently, in the VN-EdgeConv layer, k-nearest neighbor grouping degenerates into self-connections, producing edge features that are close to zero. Without a learnable bias, the resulting local features remain near zero. This issue is further exacerbated by average pooling, which propagates these near-zero features into the global representation. As a result, the network struggles to infer a meaningful global context or reconstruct the skeleton of missing regions. Moreover, because the Transformer operates on anchor features lacking discriminative activations, and no bias exists to introduce variation, it fails to recover useful information. Together, these factors prevent VN-based models from generating meaningful completions under extremely sparse observations. 

To address this limitation, we propose a rotation-equivariant bias formulation that enables the VN network to learn an initial activation shift while preserving equivariance. As described in Section 3.1, for a VN feature $\mathbf{X} \in \mathbb{R}^{C \times 3}$, we define the VN-Bias as:
\begin{equation}
    \text{VN-Bias}(\mathbf{X}; \mathbf{W}_b, \mathbf{B}) = 
    \mathbf{B} \frac{\mathbf{W}_b \mathbf{X}}{\lVert \mathbf{W}_b \mathbf{X} \rVert _F}\text{,}
    \label{eq:supp_bias}
\end{equation}
where $\mathbf{B} \in \mathbb{R}^{C^\prime \times 3}$ is the bias matrix, and $\mathbf{W}_b \in \mathbb{R}^{3 \times C}$ maps the input to a $3 \times 3$ equivariant transformation matrix. The transformation matrix is normalized by its Frobenius norm to prevent degenerate scaling. The equivariance of this bias formulation can be shown as:
\begin{equation}
    \begin{split}
        \text{VN-Bias} & (\mathbf{X}\mathbf{R}; \mathbf{W}_b, \mathbf{B}) = \mathbf{B} \frac{\mathbf{W}_b \mathbf{X} \mathbf{R}}{\lVert \mathbf{W}_b \mathbf{X} \mathbf{R} \rVert _F} \\
        & = \mathbf{B} \frac{\mathbf{W}_b \mathbf{X} \mathbf{R}} {\sqrt{\operatorname{tr}\!\left((\mathbf{W}_b \mathbf{X} \mathbf{R})^\top (\mathbf{W}_b \mathbf{X} \mathbf{R})\right)}} \\
        &= \mathbf{B} \frac{\mathbf{W}_b \mathbf{X} \mathbf{R}} {\sqrt{\operatorname{tr}\!\left(\mathbf{R}^\top \mathbf{X}^\top \mathbf{W}_b^\top \mathbf{W}_b \mathbf{X} \mathbf{R}\right)}} \\
        &= \mathbf{B} \frac{\mathbf{W}_b \mathbf{X} \mathbf{R}} {\sqrt{\operatorname{tr}\!\left(\mathbf{R}\mathbf{R}^\top \mathbf{X}^\top \mathbf{W}_b^\top \mathbf{W}_b \mathbf{X}\right)}} \\
        & = \mathbf{B} \frac{\mathbf{W}_b \mathbf{X} \mathbf{R}} {\sqrt{\operatorname{tr}\!\left((\mathbf{W}_b \mathbf{X})^\top (\mathbf{W}_b \mathbf{X})\right)}} \\
        &= \mathbf{B} \frac{\mathbf{W}_b \mathbf{X} \mathbf{R}} {\lVert \mathbf{W}_b \mathbf{X} \rVert _F} \\
        &= \text{VN-Bias}(\mathbf{X}; \mathbf{W}_b, \mathbf{B}) \mathbf{R}\text{,}
    \end{split}
    \label{eq:supp_bias_equiv}
\end{equation}
where the fourth equality follows from the cyclic property of the trace. This confirms that the proposed bias formulation preserves rotation equivariance. 

\subsection{VN-ZCALayerNorm}
As mentioned in Section 3.1 of the main paper, the VN-BatchNorm does not introduce the desired performance improvement. Therefore, we propose the VN-ZCALayerNorm and use it for all normalization layers in our framework. Zero-phase component analysis (ZCA) is a data whitening technique that transforms data to have uncorrelated components with unit variance, while preserving the original spatial structure as much as possible. 

The VN-ZCALayerNorm is formulated as:
\begin{equation}
    \mathbf{X}_i^\prime = (\mathbf{X}_i - \mu)\mathbf{W}_{\text{ZCA}} \odot \alpha\text{.}
    \label{eq:supp_zcaln}
\end{equation}
If the input is rotated to $\mathcal{X}_R = \mathcal{X}\mathbf{R}$, the mean and covariance matrix become 
\begin{equation}
    \mu_R = \mu \mathbf{R}\text{~~and~~}\Sigma_R = \mathbf{R}^\top \Sigma \mathbf{R}\text{.}
    \label{eq:supp_mu_var}
\end{equation}
The eigenvalue decomposition of $\Sigma_R$ produces 
\begin{equation}
    \Sigma_R = \mathbf{U}_R\Lambda\mathbf{U}_R^\top = \mathbf{R}^\top \mathbf{U}\Lambda\mathbf{U}^\top \mathbf{R}\text{.}
    \label{eq:supp_evd}
\end{equation}
The whitening transformation matrix is then 
\begin{equation}
    \mathbf{W}_{\text{ZCA}, R} = \mathbf{R}^\top \mathbf{U}\Lambda^{-1/2}\mathbf{U}^\top \mathbf{R} = \mathbf{R}^\top \mathbf{W}_{\text{ZCA}} \mathbf{R}\text{.}
    \label{eq:supp_wzca}
\end{equation}
We show the equivariance of the ZCA-based layer normalization as:
\begin{equation}
    \begin{split}
        \mathbf{X}_{i, R}^\prime & = (\mathbf{X}_{i, R} - \mu_R)\mathbf{W}_{\text{ZCA}, R} \odot \alpha \\
        & = (\mathbf{X}_{i}\mathbf{R} - \mu\mathbf{R})\mathbf{R}^\top\mathbf{W}_{\text{ZCA}}\mathbf{R} \odot \alpha \\
        & = (\mathbf{X}_{i} - \mu)\mathbf{R}\mathbf{R}^\top\mathbf{W}_{\text{ZCA}}\mathbf{R} \odot \alpha \\
        & = (\mathbf{X}_{i} - \mu)\mathbf{W}_{\text{ZCA}}\mathbf{R} \odot \alpha \\
        & = \mathbf{X}_{i}^\prime\mathbf{R}\text{,}
    \end{split}
\label{eq:supp_zca_equiv}
\end{equation}
where the channel-wise scaling operation $\odot$ does not affect the equivariance. It is noteworthy that theoretically, we may add a bias term $\beta \in \mathbb{R}^{C \times 3}$ here as:
\begin{equation}
    \mathbf{X}_{i}^\prime = (\mathbf{X}_{i} - \mu)\mathbf{W}_{\text{ZCA}} \odot \alpha + \beta \mathbf{U}^\top\text{,}
\label{eq:supp_zca_bias}
\end{equation}
which should not violate the rotation equivariance. However, the solution for $\mathbf{U}$ is not unique as the sign of eigenvectors could be arbitrary. This non-uniqueness is compensated in $\mathbf{W}_{\text{ZCA}}$ by multiply $\mathbf{U}$ with its transpose. But it cannot be compensated in the bias term. Thus, $\beta \mathbf{U}_R = \beta \mathbf{U}\mathbf{R}$ does not always hold, and we have to omit this bias term in VN-ZCALayerNorm.

\subsection{Discussion}

In conventional neural networks, max pooling is also an essential component as it helps the network to focus on the most activated features, making the resulting local feature more discriminative. For example, DGCNN~\cite{3d_point:dgcnn} employs max pooling within local EdgeConv to enhance discriminability. However, due to the non-differentiable selection mechanism in VN-MaxPooling, it cannot be effectively trained. As a result, VN-EdgeConv often resorts to average pooling, which may reduce its expressive power. This limitation highlights the need for future work on incorporating effective max pooling mechanisms into VN layers, without compromising rotation equivariance, to further improve the representational capacity of VN-based models.

\section{Experiments}

\subsection{More Ablation Study}

In addition to the module design, we conduct an ablation study for different module hyperparameters. Table~\ref{tab:supp_ablation_layers} evaluates the impact of the depth in the VN Missing Anchor Transformer (VN-MATr), where $N_\text{enc}$ and $N_\text{dec}$ denote the number of encoder blocks and decoder blocks, respectively. We observe that increasing the Transformer depth consistently contributes to the model accuracy. However, the improvement of the configuration $(N_\text{enc}=6, N_\text{dec}=8)$ over $(N_\text{enc}=4, N_\text{dec}=6)$ is marginal. Therefore, we decide to apply $(N_\text{enc}=4, N_\text{dec}=6)$ as our final Transformer depth, which is the largest possible configuration for a single NVIDIA RTX 5090 GPU.

\setlength{\tabcolsep}{5pt}
\begin{table}
    \centering
    \small
    \begin{tabular}{c c|c c c}
        \hline
        $N_\text{enc}$ & $N_\text{dec}$ & CD-P$\downarrow$ & $\text{F1}@1\%\uparrow$ & $\text{F1}@2\%\uparrow$ \\
        \hline
        0 & 2 & 0.931 & 70.97 & 91.30 \\
        2 & 3 & 0.907 & 71.92 & 91.72 \\
        3 & 4 & 0.901 & 72.86 & 92.06 \\
        4 & 6 & 0.887 & 73.44 & 92.37 \\
        6 & 8 & 0.885 & 73.50 & 92.48 \\
        \hline
    \end{tabular}
    \caption{Ablation study of the depth of VN-MATr layers on the MVP dataset using the resolution of 8192 points.}
    \label{tab:supp_ablation_layers}
\end{table}

\subsection{More Quantitative Results}

Tables~\ref{tab:supp_mvp_f1_001} and~\ref{tab:supp_mvp_f1_002} provide the detailed F-Score results on the MVP dataset. Our model achieves the highest F-Score across nearly all object categories, and the substantial gains under the stricter F-Score@1\% metric further indicate its superior ability to recover fine geometric details.

\setlength{\tabcolsep}{3pt}
\begin{table}
    \centering
    \small
    \begin{tabular}{c|c c c c c c c c|c}
        \hline
        & \multicolumn{9}{c}{F-Score@1\%$\uparrow$} \\
        \hline
        \hline
        Method & Air & Cab & Car & Cha & Lam & Sof & Tab & Ves & Avg\\
        \hline
        & \multicolumn{9}{c}{SO(3)/SO(3)}\\
        \hline
        PoinTr & 74.34 & 43.48 & 49.85 & 54.20 & 64.89 & 48.10 & 57.44 & 61.21 & 56.69 \\
        AnchorFormer & 78.01 & 40.84 & 45.99 & 52.99 & 64.48 & 46.43 & 57.78 & 60.13 & 55.83 \\
        PMP-Net++ & 80.52 & 46.06 & 48.74 & 59.11 & 72.39 & 50.54 & 62.97 & 63.39 & 60.47 \\
        GTNet & 83.12 & 47.12 & 50.89 & 58.78 & 71.25 & 51.13 & 61.83 & 65.28 & 61.17 \\
        ODGNet & 88.92 & 53.72 & 57.74 & 65.64 & 78.52 & 57.71 & 68.95 & 69.63 & 67.61 \\
        AdaPoinTr & 90.15 & 54.89 & 57.96 & 64.30 & 78.03 & 57.35 & 69.74 & 71.71 & 68.02 \\
        \hline
        & \multicolumn{9}{c}{None/SO(3)}\\
        \hline
        EquivPCN & 89.36 & 54.94 & 58.54 & 64.74 & 75.45 & 57.48 & 69.68 & 69.99 & 67.52 \\
        ESCAPE & 81.11 & 47.00 & 51.59 & 59.74 & 69.66 & 50.13 & 63.91 & 61.74 & 60.61 \\
        Ours & \textbf{91.89} & \textbf{62.70} & \textbf{65.53} & \textbf{70.61} & \textbf{80.55} & \textbf{64.91} & \textbf{76.96} & \textbf{74.35} & \textbf{73.44} \\
        \hline
    \end{tabular}
    \caption{Point cloud completion accuracy in terms of F-Score@1\% $\uparrow$ on the MVP dataset using the resolution of 8192 points. The best results are highlighted bold. Our model achieves the best F-Score@1\% in all categories. }
    \label{tab:supp_mvp_f1_001}
\end{table}

\setlength{\tabcolsep}{3pt}
\begin{table}
    \centering
    \small
    \begin{tabular}{c|c c c c c c c c|c}
        \hline
        & \multicolumn{9}{c}{F-Score@2\%$\uparrow$} \\
        \hline
        \hline
        Method & Air & Cab & Car & Cha & Lam & Sof & Tab & Ves & Avg\\
        \hline
        & \multicolumn{9}{c}{SO(3)/SO(3)}\\
        \hline
        PCN & 90.59 & 73.22 & 80.48 & 69.69 & 78.67 & 72.00 & 74.08 & 83.85 & 77.82 \\
        TopNet & 77.47 & 64.33 & 69.18 & 58.22 & 72.23 & 63.79 & 58.63 & 73.71 & 67.20 \\
        MSN & 89.32 & 72.23 & 78.76 & 77.66 & 82.98 & 71.69 & 78.92 & 81.58 & 79.14 \\
        CRN & 91.40 & 73.70 & 81.36 & 76.16 & 79.65 & 73.24 & 78.31 & 83.87 & 79.71 \\
        GRNet & 90.92 & 73.63 & 81.61 & 79.39 & 84.53 & 74.81 & 83.17 & 83.95 & 81.50 \\
        VRCNet & 95.65 & 81.11 & 87.49 & 82.45 & 88.45 & 80.72 & 86.23 & 88.15 & 86.28 \\
        PoinTr & 92.42 & 77.18 & 83.34 & 77.99 & 84.97 & 76.43 & 82.08 & 86.25 & 82.58 \\
        AnchorFormer & 94.19 & 74.91 & 81.93 & 77.95 & 84.87 & 74.34 & 84.45 & 86.84 & 82.44 \\
        PMP-Net++ & 96.00 & 71.16 & 75.65 & 79.42 & 87.71 & 72.68 & 84.52 & 85.82 & 81.62 \\
        GTNet & 96.10 & 79.36 & 83.80 & 82.37 & 88.37 & 79.35 & 86.53 & 88.38 & 85.53 \\
        ODGNet & 97.27 & 84.73 & 89.89 & 86.71 & 91.95 & 84.73 & 89.74 & 91.11 & 89.52 \\
        AdaPoinTr & 97.90 & 86.79 & 90.16 & 88.50 & 92.51 & 86.14 & 91.92 & \textbf{92.51} & 90.80 \\
        \hline
        & \multicolumn{9}{c}{None/SO(3)}\\
        \hline
        EquivPCN & 96.69 & 82.14 & 87.18 & 82.25 & 87.14 & 80.79 & 85.74 & 88.70 & 86.33 \\
        ESCAPE & 93.66 & 81.81 & 86.19 & 85.05 & 87.53 & 80.56 & 87.91 & 84.36 & 85.88 \\
        Ours & \textbf{98.00} & \textbf{90.07} & \textbf{92.87} & \textbf{90.60} & \textbf{93.04} & \textbf{89.03} & \textbf{93.24} & 92.10 & \textbf{92.37} \\
        \hline
    \end{tabular}
    \caption{Point cloud completion accuracy in terms of F-Score@2\% $\uparrow$ on the MVP dataset using the resolution of 8192 points. The best results are highlighted bold. Our model achieves the best F-Score@2\% in almost all categories. }
    \label{tab:supp_mvp_f1_002}
\end{table}

\subsection{Failure Cases on MVP}
\begin{figure}
    \centering
    \scriptsize
    \renewcommand{\arraystretch}{0.5}
    \begin{tabular}{m{30pt}<{\centering} m{30pt}<{\centering} m{30pt}<{\centering} m{30pt}<{\centering} m{30pt}<{\centering} m{30pt}<{\centering} m{30pt}<{\centering} m{30pt}<{\centering} m{30pt}<{\centering} m{30pt}<{\centering}}
        Input & Anchor-Former & PMP-Net++ & GTNet & ODG-Net & Ada-PoinTr & Equiv-PCN & ESCAPE & Ours & Ground-truth\\

        \includegraphics[width=30pt]{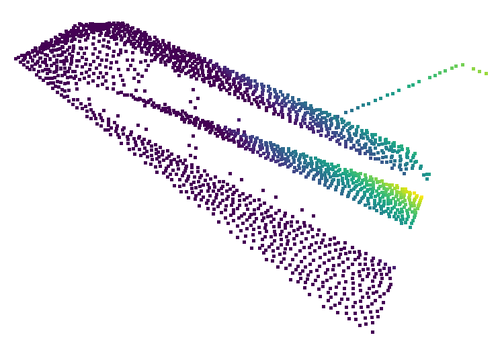} & 
        \includegraphics[width=30pt]{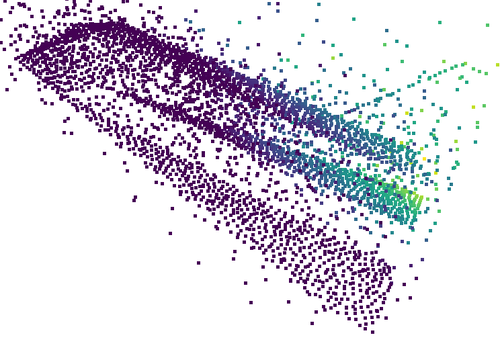} & 
        \includegraphics[width=30pt]{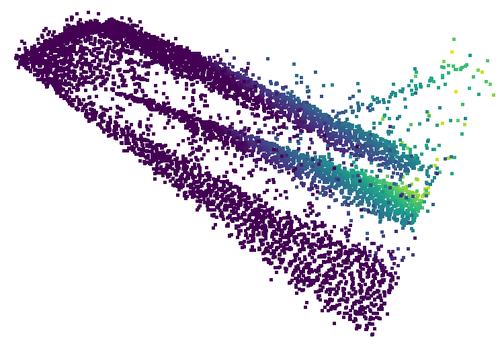} & 
        \includegraphics[width=30pt]{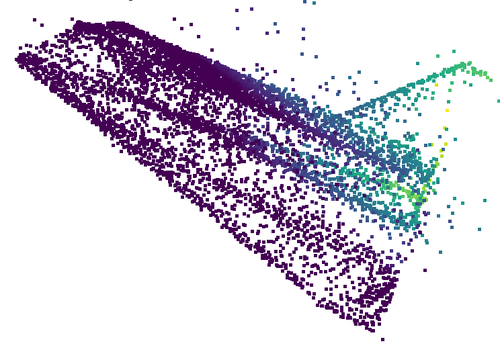} & 
        \includegraphics[width=30pt]{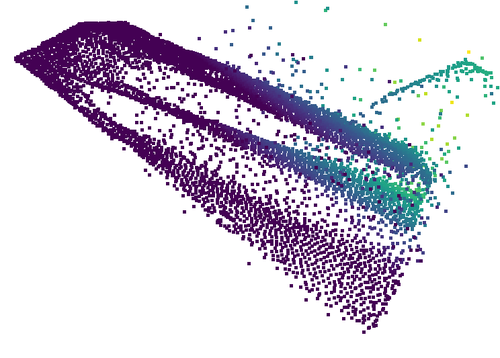} & 
        \includegraphics[width=30pt]{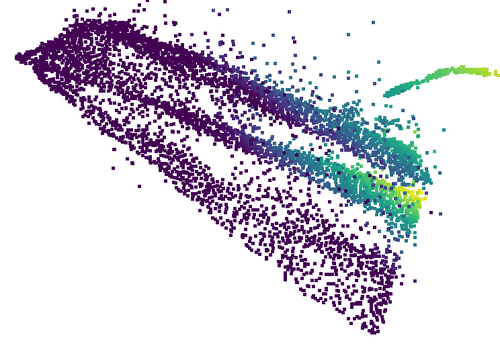} & 
        \includegraphics[width=30pt]{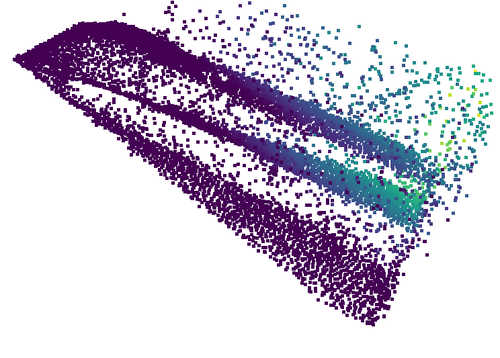} & 
        \includegraphics[width=30pt]{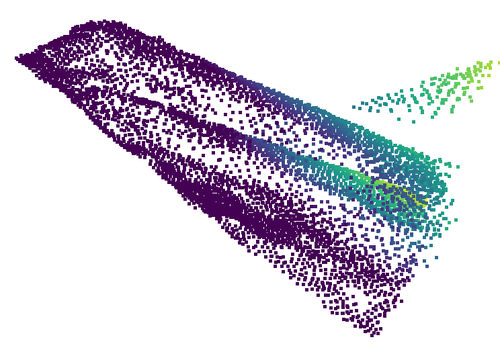} & 
        \includegraphics[width=30pt]{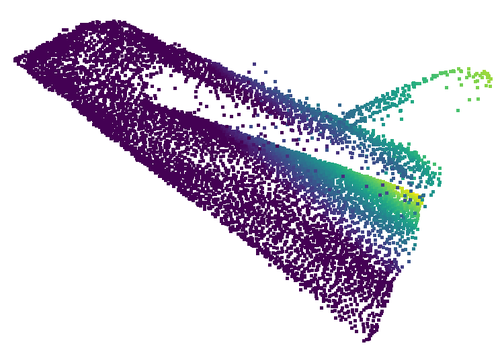} & 
        \includegraphics[width=30pt]{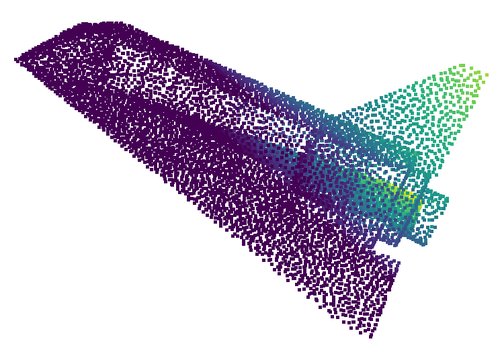} \\

        \includegraphics[width=30pt]{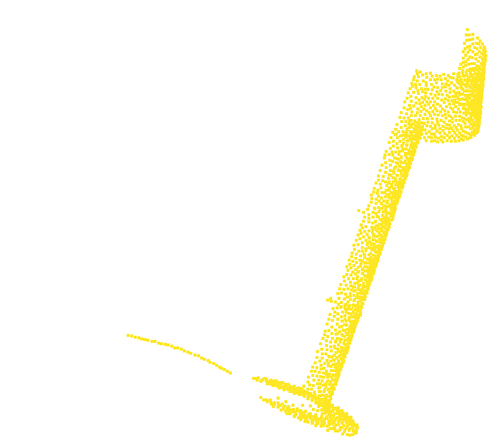} & 
        \includegraphics[width=30pt]{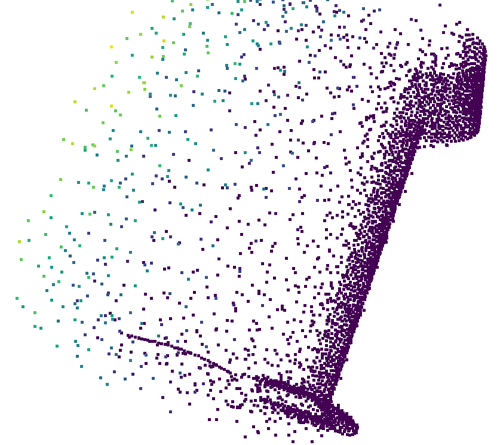} & 
        \includegraphics[width=30pt]{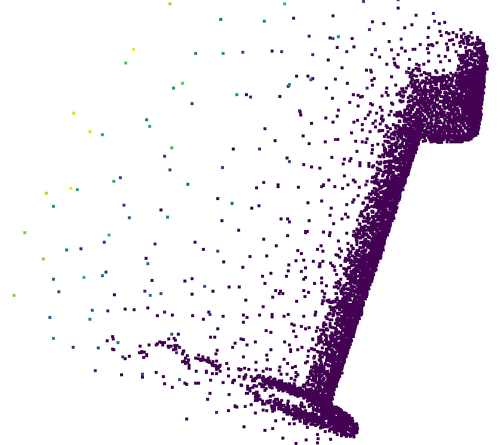} & 
        \includegraphics[width=30pt]{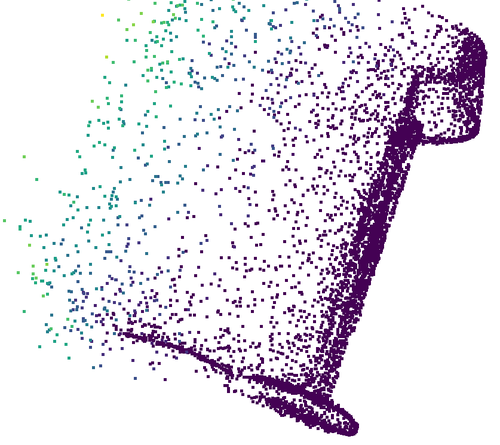} & 
        \includegraphics[width=30pt]{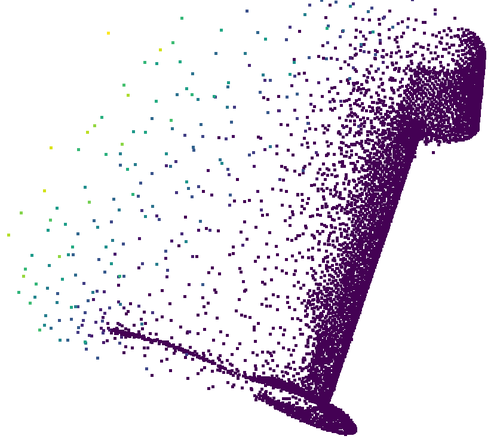} & 
        \includegraphics[width=30pt]{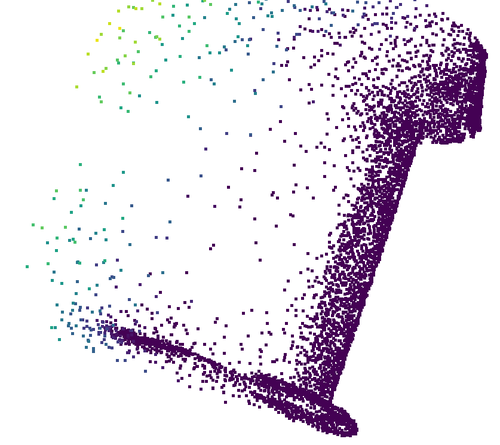} & 
        \includegraphics[width=30pt]{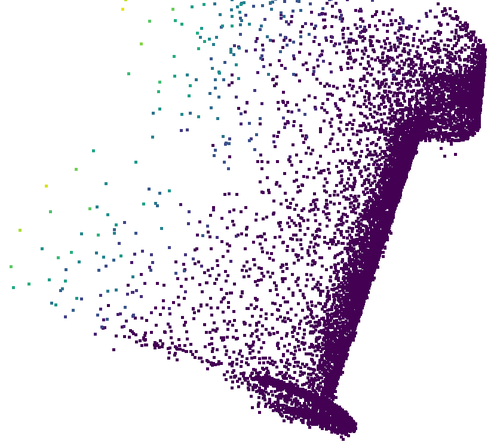} & 
        \includegraphics[width=30pt]{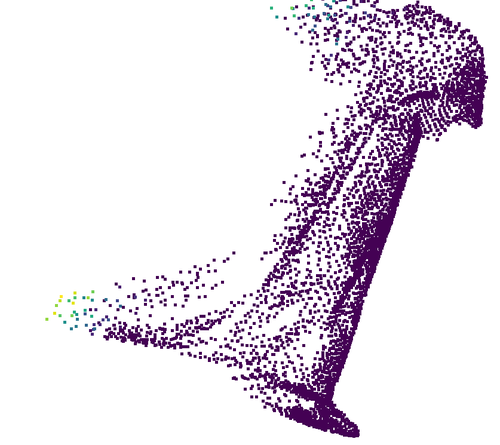} & 
        \includegraphics[width=30pt]{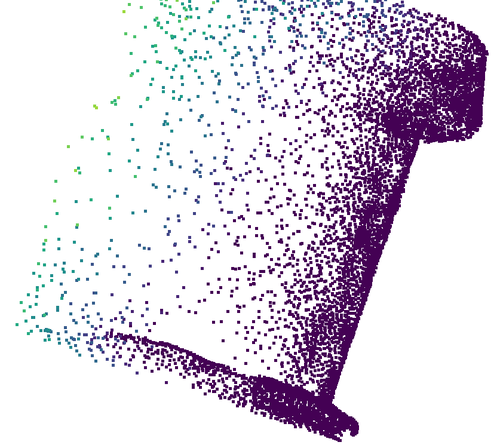} & 
        \includegraphics[width=30pt]{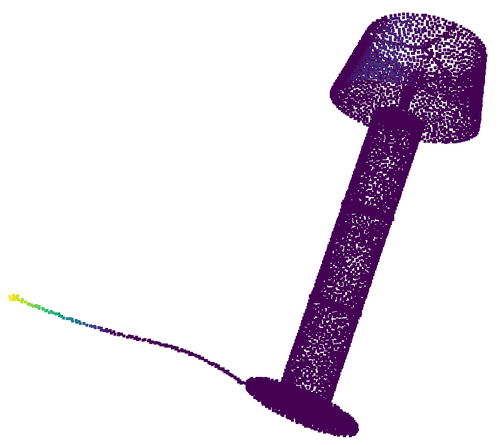} \\

        \includegraphics[width=30pt]{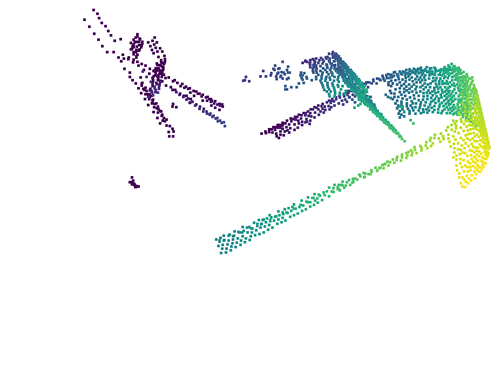} & 
        \includegraphics[width=30pt]{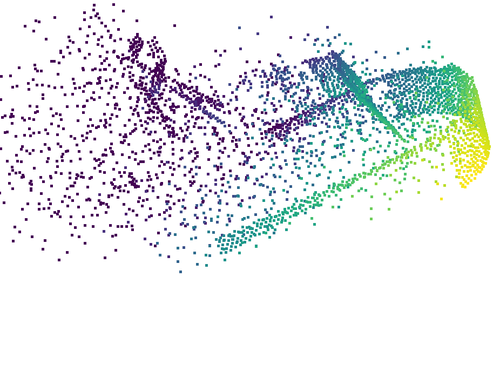} & 
        \includegraphics[width=30pt]{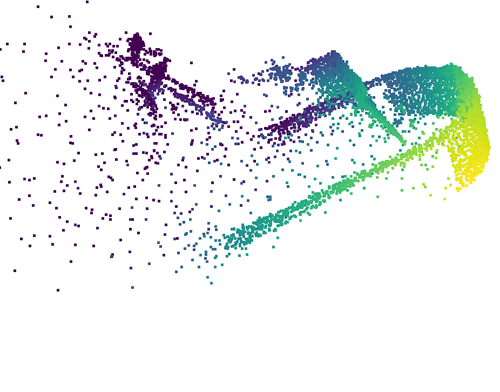} & 
        \includegraphics[width=30pt]{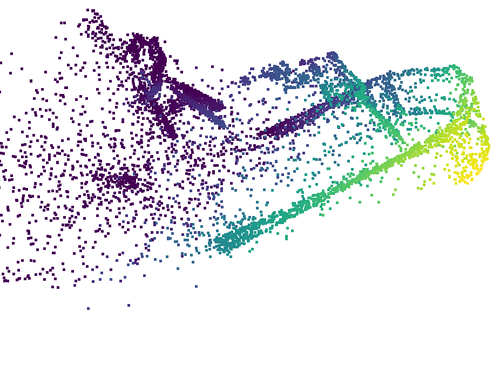} & 
        \includegraphics[width=30pt]{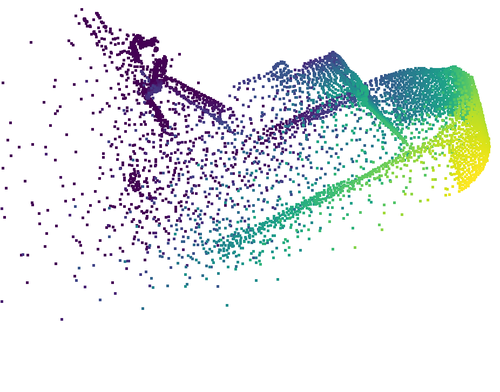} & 
        \includegraphics[width=30pt]{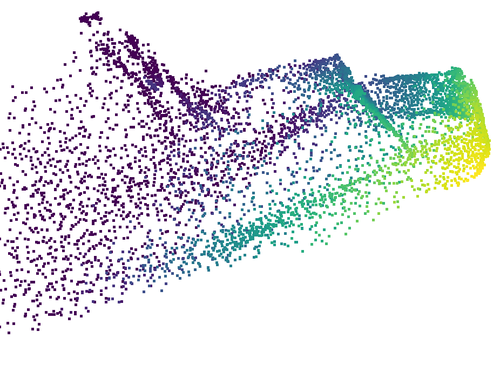} & 
        \includegraphics[width=30pt]{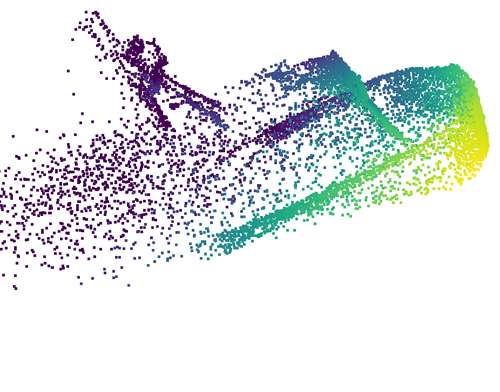} & 
        \includegraphics[width=30pt]{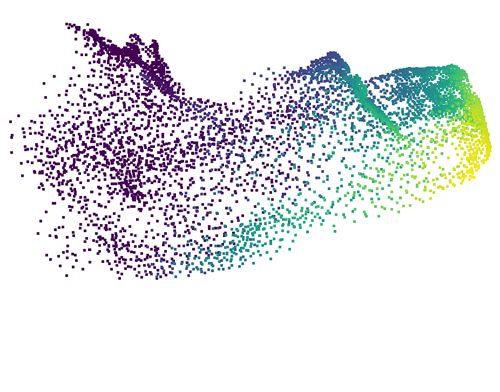} & 
        \includegraphics[width=30pt]{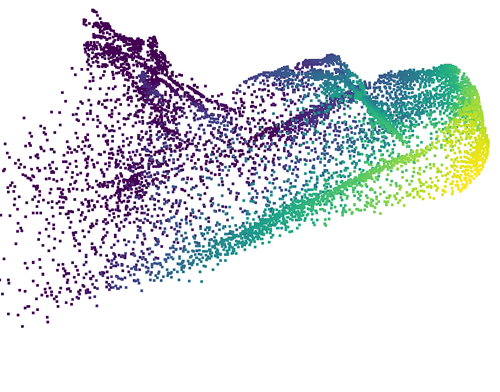} & 
        \includegraphics[width=30pt]{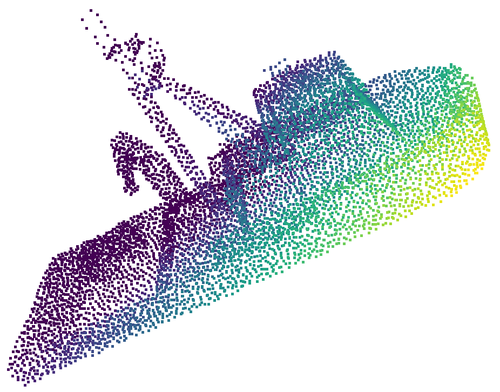} \\
    \end{tabular}
    \caption{Failure cases on MVP dataset.}
    \label{fig:supp_failure_mvp}
\end{figure}

We illustrate several failure cases on the MVP dataset in Fig.~\ref{fig:supp_failure_mvp}. A common cause of failure lies in the model's difficulty in inferring the underlying skeleton of the missing regions based on the partial observation. For instance, our model struggles with the following objects: (1) An airplane, where certain unusual shapes are underrepresented in the training data; (2) A lamp, where the model fails to understand the cable structure; (3) A vessel, which has complex and non-repetitive geometry. These failures largely stem from the lack of structural priors for thin parts such as lamp cables or vessel masts, and the difficulty of modeling highly complex or asymmetric shapes like vessels. Future work may address these challenges by incorporating symmetry-aware priors and enriching the training set with more diverse and long-tail examples. For the lamp example, we further observe that the presence of cables shifts the lamp body away from the input center, which may significantly impact the performance of all models. This highlights the need for future work to also consider translation invariance, in addition to rotation, for more robust completion.

\setlength{\tabcolsep}{1.5pt}

\end{document}